\documentclass[11pt]{article}

\usepackage[preprint]{acl}
\usepackage{times}
\usepackage{latexsym}

\usepackage[T1]{fontenc}
\usepackage[utf8]{inputenc}

\usepackage{microtype}

\usepackage{inconsolata}
\usepackage{enumitem,comment}
\usepackage{graphicx}
\usepackage{amsmath}
\usepackage{booktabs}
\usepackage{pifont}
\usepackage{makecell} 
\usepackage{float}

\definecolor{mygreen}{RGB}{0,128,0}
\definecolor{myred}{RGB}{180,0,0}

\providecommand{\cmark}{\textcolor{mygreen}{\ding{51}}}
\providecommand{\xmark}{\textcolor{myred}{\ding{55}}}

\usepackage[table]{xcolor}
\definecolor{deepred}{HTML}{F88379}
\definecolor{lightred}{HTML}{FFC0CB}
\usepackage{multirow}

\title{LLMRouterBench: A Massive Benchmark and Unified Framework for LLM Routing}

\author{Hao Li\thanks{Equal contribution.}\footnotemark[3]\footnotemark[4] \quad Yiqun Zhang\footnotemark[1]\footnotemark[4]  \quad Zhaoyan Guo\footnotemark[1]\footnotemark[3] \quad Chenxu Wang\footnotemark[1]\footnotemark[4] \\
\textbf{\quad Shengji Tang\quad Qiaosheng Zhang\quad Yang Chen \quad Biqing Qi} \\ \textbf{Peng Ye \quad Lei Bai \quad Zhen Wang\thanks{Correspondence: Zhen Wang (w-zhen@nwpu.edu.cn); Shuyue Hu (hushuyue@pjlab.org.cn).}\footnotemark[3]\quad Shuyue Hu\footnotemark[2]}\\
\\
Shanghai Artificial Intelligence Laboratory
} 

\begin{document}
\maketitle
\renewcommand{\thefootnote}{\fnsymbol{footnote}}
\footnotetext[3]{Hao Li, Zhaoyan Guo, and Zhen Wang are affiliated with Northwestern Polytechnical University.}
\footnotetext[4]{This work was done during their internship at Shanghai Artificial Intelligence Laboratory.}

\begin{abstract}
Large language model (LLM) routing assigns each query to the most suitable model from an ensemble. We introduce LLMRouterBench, a large-scale benchmark and unified framework for LLM routing. It comprises over 400K instances from 21 datasets and 33 models. Moreover, it provides comprehensive metrics for both performance-oriented and performance-cost trade-off routing, and integrates 10 representative routing baselines. Using LLMRouterBench, we systematically re-evaluate the field. While confirming strong model complementarity—the central premise of LLM routing—we find that many routing methods exhibit similar performance under unified evaluation, and several recent approaches, including commercial routers, fail to reliably outperform a simple baseline. Meanwhile, a substantial gap remains to the Oracle, driven primarily by persistent model-recall failures. We further show that backbone embedding models have limited impact, that larger ensembles exhibit diminishing returns compared to careful model curation, and that the benchmark also enables latency-aware analysis. All code and data are available at \url{https://github.com/ynulihao/LLMRouterBench}.
\end{abstract}

\renewcommand{\thefootnote}{\arabic{footnote}}
\setcounter{footnote}{0}

\section{Introduction}
\begin{figure}[t]
    \centering   
    \includegraphics[width=0.94\linewidth]{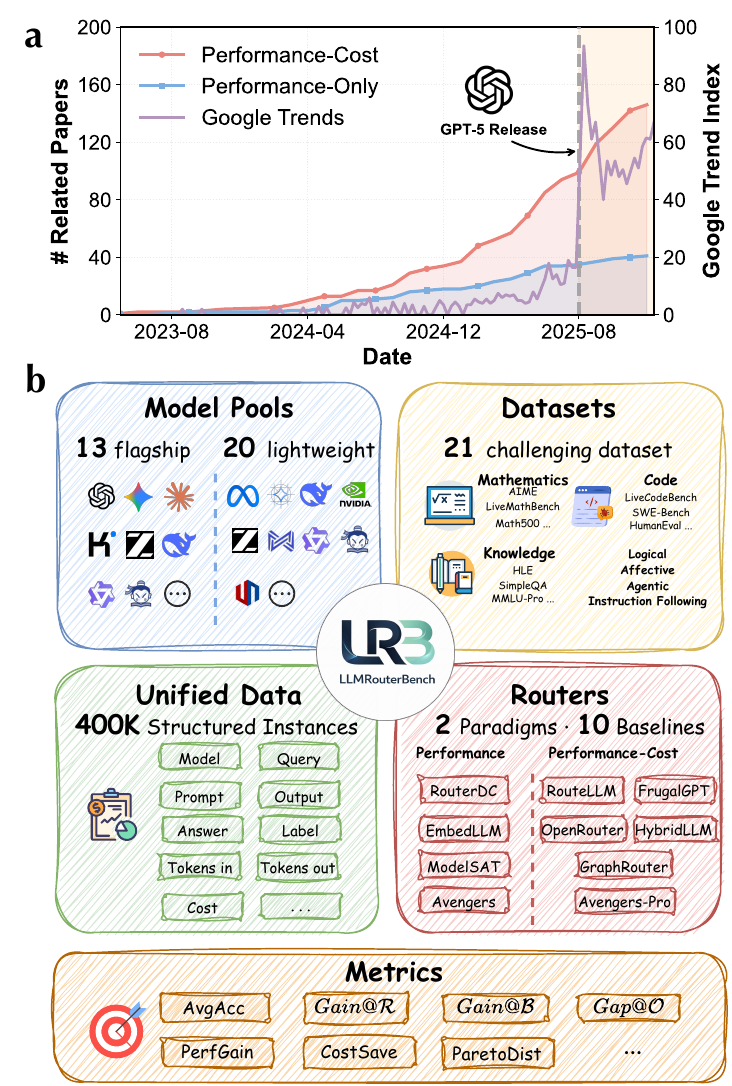}
    % \vspace{-1.6em}
    \caption{(a) Interest in LLM routing over time, as measured by the cumulative number of related papers and Google Trends. (b) An Overview of LLMRouterBench.
    }
    \vspace{-1.8em}
    \label{fig:openrouterbench-overview}
\end{figure}

% \footnote{For example, as of January 2026, there are over 2,165 models fine-tuned on LLaMA-3.1-8B-Instruct and 778 on Qwen-3-8B.}
The rapid evolution of large language models (LLMs) has led to a proliferation of publicly available models. In this landscape, LLM routing has emerged as an important direction: rather than relying on a single model, routing methods operate over an ensemble of LLMs and dynamically assign each query to the model best suited to handle it. Since early studies in 2023 that focused primarily on improving the collective performance of LLM ensembles~\cite{jiang-etal-2023-llm,lu-etal-2024-routing,huang2025lookahead}, the field has broadened to address performance-cost trade-offs~\cite{ding2024hybrid,wang2025mixllm,ding2025bestroute,jitkrittum2025universal};  more recently, the integration of real-time routing into production systems such as GPT-5~\cite{openai_gpt5_2025} and HuggingChat-Omni~\cite{huggingface_omni_2025} has further boosted academic, industrial, and even public interest (Fig.~\ref{fig:openrouterbench-overview} (a)).

Despite its academic origins, progress in this field risks becoming increasingly marginalized within the research community for two primary reasons. First, developing a routing method typically requires evaluating an ensemble of LLMs across a wide range of datasets; this incurs \textit{substantial computational costs} from deploying multiple large models on dedicated hardware or \textit{significant financial costs} from relying on online service providers (e.g., OpenRouter). Second, the \textit{lack of open-source infrastructure} has led to a highly fragmented research landscape: individual studies implement bespoke routing pipelines using different model ensembles, datasets, and evaluation protocols, hindering fair comparison, reproducibility, and cumulative progress in the field.

To reduce these barriers for the community, we introduce \emph{LLMRouterBench}, a benchmark and framework for LLM routing Fig.~\ref{fig:openrouterbench-overview} (b). As summarized in Table 1, unlike existing benchmarks that rely largely on earlier-generation open-source models or evaluate only a narrow class of routing methods, LLMRouterBench provides \textit{timely}, \textit{large-scale}, \textit{high-quality} data curated from 21 datasets and 33 models. These include 13 recently released flagship models and 5 widely used proprietary models, totaling over 400K instances, \$2.7K in API costs, and approximately 1K GPU hours. Beyond scale, LLMRouterBench natively supports both dominant research paradigms in this field (i.e. performance-oriented routing and performance-cost tradeoff routing), defines a comprehensive suite of metrics for each setting, and offers adapters to interface with publicly available routing implementations with minimal modification; crucially, it enables \textit{reproducible}, \textit{unified} evaluation across 10 representative routing baselines.

With LLMRouterBench, we systematically re-examine the landscape of LLM routing and uncover several key findings. First, we reaffirm the field’s central premise: models exhibit clear \textit{complementarity} both in performance and cost-efficiency, with \emph{no} single model achieving universal dominance. Second, by evaluating routing baselines on a unified model ensemble and dataset suite, we identify a surprising pattern: despite ongoing methodological innovation, contemporary routing approaches deliver nearly \textit{indistinguishable} results in performance across various metrics. This pattern also extends to performance-cost tradeoff settings: several recent routing methods, even including the commercial router OpenRouter, do \textit{not} outperform a simple baseline (the Best Single model), \textit{nor} do they reliably reduce cost without sacrificing performance relative to Best Single. These findings suggest that under unified, large-scale evaluation, the practical gains of current routing methods may be \textit{less} significant  than previously assumed. 

Third, although the above results may initially appear discouraging, we find that LLM routing remains far from its capability ceiling: a substantial performance gap persists relative to an Oracle baseline that always selects the best-performing model per query in hindsight. This gap is primarily driven by persistent \textit{model-recall failures}: for many queries, only a single candidate model produces a correct response, yet current routers frequently fail to identify it, highlighting a clear opportunity for future improvement. Fourth, while several routing methods rely on embedding models, our ablation study shows that embeddings have \textit{little} impact on routing performance. Fifth, although expanding the model ensemble is often assumed to improve collective performance~\cite{huang2025routereval}, we observe clear \textit{diminishing} returns from adding more models; in contrast, a carefully selected subset can yield substantially better outcomes. This suggests that model curation should be studied jointly with routing, as careful curation can outweigh simply scaling the ensemble. Finally, we show that LLMRouterBench also enables latency-aware analysis, opening a path toward performance-cost-latency optimization in future routing research.

Our key contributions are summarized as follows:
\begin{itemize}[leftmargin=*,itemsep=0.15em, topsep=0.2em]
   \item \textbf{Timely, high-quality, large-scale data:} 400K+ instances curated from 21 challenging datasets and 33 recently released models, lowering the cost barrier and improving the accessibility of LLM routing research;
    \vspace{-0.5em}
    \item \textbf{An open-source, unified routing framework:} integration of 10 representative routing baselines and comprehensive quantitative metrics in both performance-oriented and performance-cost tradeoff settings, enabling fair, reproducible evaluation;
    \vspace{-0.5em}
    \item \textbf{A systematic re-examination of LLM routing:} an extensive comparative study of leading routing methods that systematically analyzes their strengths and limitations, characterizes the capability ceiling, rigorously tests common practices and prevailing hypotheses, and outlines promising directions for future research.
\end{itemize}

% \begin{figure}[t]
%     \centering   
%     \includegraphics[width=0.95\linewidth]{figs/figure1-overview.pdf}
%     \caption{An Overview of LLMRouterBench.}
%     \label{fig:openrouterbench-overview}
%     \vspace{-1em}
% \end{figure}

\begin{table*}[!htbp]
  \centering
  \small
  \resizebox{\textwidth}{!}{%
  \setlength{\tabcolsep}{1.5pt}
  % ~\cite{hu2024routerbench}
  % ~\cite{zhuang2024embedllm}
  % ~\cite{huang2025routereval}
  % ~\cite{feng2025fusionfactory}
  % ~\cite{lu2025routerarena}
  \begin{tabular}{@{}l c c c c c c c@{}}
    \toprule
    \textbf{Benchmark}
    & \textbf{Performance-oriented}
    & \textbf{Performance--Cost}
    & \textbf{Datasets}
    & \makecell[c]{\textbf{Proprietary}\\\textbf{Models}}
    & \makecell[c]{\textbf{Unified}\\\textbf{Model Pool}}
    & \makecell[c]{\textbf{Routing}\\\textbf{Baselines}}
    & \makecell[c]{\textbf{Data}\\\textbf{Release}} \\
    \midrule
    RouterBench   & \xmark & \cmark & 8  & \makecell[c]{GPT-4\&3.5, Claude v1\&2} & \cmark & 3 & \cmark \\
    EmbedLLM      & \cmark & \xmark & 10 & \xmark & \cmark & 1 & \cmark \\
    RouterEval    & \cmark & \xmark & 12 & \xmark & \cmark  & 4 & \cmark \\
    FusionFactory & \xmark & \cmark & 14 & \xmark & \cmark & 5 & \cmark \\
    RouterArena   & \xmark & \cmark & 23\textsuperscript{†} & \cmark\textsuperscript{‡} & \xmark & \xmark & \xmark \\
    \midrule
     LLMRouterBench & \cmark & \cmark & 21 & \makecell[c]{GPT-5, GPT-5-Chat, Claude 4\\Gemini 2.5 Pro...} & \cmark & 10 & \cmark \\
    \bottomrule
  \end{tabular}
  }
  \caption{Comparison with existing routing benchmarks.
  \textsuperscript{†}RouterArena treats the routing system as a black box, reporting only overall performance without per-prompt, per-model details, and thus \textbf{cannot} directly support developing new routing algorithms.
  \textsuperscript{‡}Each routing system uses different models.}
  \vspace{-1.5em}
  \label{tab:router-benchmarks}
\end{table*}

\section{Related Work}
\paragraph{Routing for Performance.}
This paradigm aims to enhance collective performance of an ensemble of LLMs by routing each query to the model that is most likely to produce a correct answer~\cite{yue2025masrouter,zhang2025router,fein2025mixture,wang2025icl}.
LLM-Blender~\cite{jiang-etal-2023-llm} selects top candidate models through pairwise ranking and fuses their outputs.
% ZOOTERS~\cite{lu-etal-2024-routing} utilizes reward-guided routing enhanced by tag-based supervision.
RouterDC~\cite{chen2024routerdc} applies dual contrastive learning to boost routing accuracy.
EmbedLLM~\cite{zhuang2024embedllm} leverages compact model and query embeddings.
MODEL-SAT~\cite{zhang2025capability} constructs capability-aligned embeddings to train a lightweight LLM as a router.
% Router-R1~\cite{zhang2025router} models multi-hop QA routing as a sequential decision process with reinforcement learning.
% MasRouter~\cite{yue2025masrouter} generalizes routing from single LLMs to multi-agent systems and employs a cascaded controller for coordination mechanism selection.
% ICL-Router~\cite{wang2025icl} represents model capabilities with in-context vectors, enabling seamless integration of new LLMs without retraining.
% Lookahead~\cite{huang2025lookahead} predicts candidate model outputs in latent space to enable response-aware routing without full inference.
Avengers~\cite{zhang2025avengers} shows that a clustering-based routing, combined with aggregation methods, can empower a set of small LLMs to surpass proprietary models.
% Recent studies further extend routing beyond single-model selection to support multi-agent coordination~\cite{yue2025masrouter} and multi-hop reasoning~\cite{zhang2025router}.

\paragraph{Routing for Performance-Cost Tradeoff.}
This paradigm routes queries to models to tradeoff between performance and cost~\cite{song-etal-2025-irt,jin-etal-2025-radialrouter,patel2025proxrouter,fernandez2025radar,guo2025towards}, as more capable models are often associated with higher financial and computational costs.
Earlier studies primarily focused on routing between only two models (typically a small one and a larger one).
HybridLLM~\cite{ding2024hybrid} targets the tradeoff based on predicted query difficulty.
FrugalGPT~\cite{chen2024frugalgpt} adopts cascaded inference from small to large models. 
RouteLLM~\cite{ong2024routellm} trains the router from preference data. 
More recently, research has extended to larger model ensembles.
Avengers-Pro~\cite{zhang2025beyond} are both build upon clustering-based routing.
GraphRouter~\cite{feng2025graphrouter} employs a heterogeneous graph to represent task-query-LLM interactions, formulating model routing as an edge prediction problem.
% BEST-Route~\cite{ding2025bestroute} routes queries between a reference LLM and best-of-n sampling on small models.
% Universal Router~\cite{jitkrittum2025universal} employs a cluster-based routing strategy that groups queries based on unlabeled data, selecting LLMs using cost-adjusted error scores computed per cluster on a labeled validation set.
% Similarly, Avengers-Pro~\cite{zhang2025beyond}, which is also built upon cluster-based routing, outperforms GPT-5 on diverse challenging tasks with significantly less inference costs.

% \paragraph{Routing in new application scenarios.}
% LLM routing has been generalized to various novel application scenarios beyond this study.  
% MasRouter~\cite{yue2025masrouter} and STRMAC~\cite{wang2025optimal} generalize routing from single LLMs to multi-agent systems. MasRouter builds query-specific workflows by jointly choosing roles, LLMs, and collaboration structures, while STRMAC learns a state-aware policy that decides which agent should act at each step. Router-R1~\cite{zhang2025router} addresses multi-hop QA by sequentially selecting appropriate models. Beyond interactive settings, routing also supports budget-aware data generation and distillation~\cite{zhang2025find} by mixing model pool without sacrificing performance.

\section{LLMRouterBench}

\begin{figure*}[t]
    \centering   
    \includegraphics[width=0.95\linewidth]{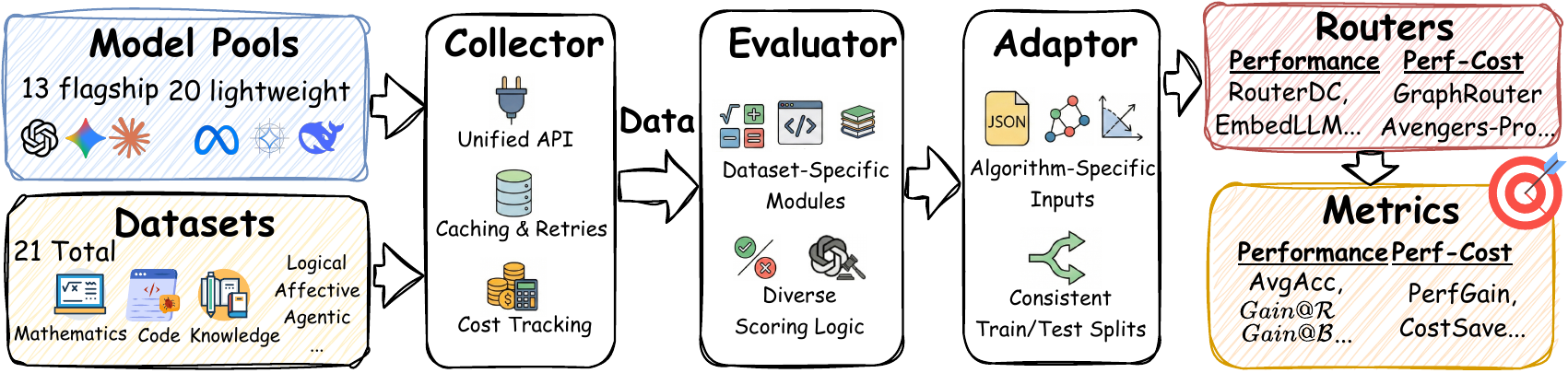}
    \caption{LLMRouterBench framework integrating Collector, Evaluator, and Adaptor components for standardized evaluation of LLM routing methods.}
    \vspace{-1.3em}
    \label{fig:framework}
\end{figure*}

We construct the LLMRouterBench to systematically support two dominant paradigms in LLM routing research: routing for \emph{performance} and routing for \textit{performance-cost tradeoff}. %Each paradigm has its own tailored model pool and dataset suite, detailed below.

\subsection{Model Pools}

In the performance-oriented setting, we benchmark routing methods' ability to exploit complementary strengths among comparable-sized LLMs.\footnote{It is trivial to consider models with significant variations in sizes, as larger models typically yield better performance than smaller models.} In contrast, the performance-cost setting intentionally includes models with substantial variations in size, capability, and cost to reflect realistic performance-cost tradeoffs. This yields two pools totaling 33 models:
\begin{itemize}[leftmargin=*,itemsep=0.15em, topsep=0.2em]
    \item For the performance-oriented setting, we construct a pool of 20 state-of-the-art $\sim$7B \textbf{lightweight LLMs}, such as DS-Qwen3 and Qwen3-8B (see Appendix Table~\ref{tab:perf-model-pool} for details)
    \item For the performance-cost setting, we build another pool of 13 \textbf{flagship LLMs} from 8 providers, varying substantially in capability and cost, such as GPT-5 and Gemini-Flash, with cost information collected from OpenRouter~\cite{openrouter} and official APIs (see Appendix Table~\ref{tab:cost-model-pool} for details).
\end{itemize}

\subsection{Datasets}

We curate 21 datasets spanning multiple domains (full list in Appendix Table~\ref{tab:benchmarks}), including \textbf{Mathematics} (e.g., AIME, LiveMathBench~\cite{liu2024livemathbench}), \textbf{Code} (e.g., HumanEval~\cite{chen2021evaluating}, SWE-Bench~\cite{jimenez2023swe}), \textbf{Logic} (e.g., BBH~\cite{suzgun2022challenging}, KORBench~\cite{ma2024korbenchbenchmarkinglanguagemodels}), \textbf{Knowledge} (e.g., HLE~\cite{phan2025humanity}, SimpleQA~\cite{wei2024measuring}), \textbf{Affective} (e.g., MELD~\cite{poria2019meldmultimodalmultipartydataset}), \textbf{Instruction Following} (e.g., ArenaHard~\cite{li2024crowdsourced}), and \textbf{Tool Use} (e.g., $\tau^2$-Bench~\cite{barres2025tau}).

% \textit{Mathematics}, \textit{Code}, \textit{Logical Reasoning}, \textit{Knowledge}, \textit{Affective Computing}, \textit{Instruction Following}, and \textit{Agentic Tasks}) with varying difficulty levels. 
For flagship models under the performance-cost setting, we select  10 datasets, excluding saturated datasets to effectively evaluate frontier capabilities. For lightweight models under the performance-oriented setting, we select 15 datasets, excluding excessively challenging datasets on which all models perform uniformly poorly to ensure meaningful comparisons (See Appendix Table ~\ref{tab:benchmarks} for details.).  %Specifically, the performance-only setting includes 15 datasets suitable for lightweight models, while the performance-cost setting includes 10 more demanding datasets tailored for flagship models 

% \vspace{-1.4em}

% \input{tables/table6-1_datasets}
% \input{tables/table6-2_datasets}

\begin{comment}
\subsection{Data Format}

LLMRouterBench  adopts a standardized, dataset-agnostic data format suitable for evaluating both performance-only and performance-cost tradeoff routing methods. Formally, each routing instance is represented by a tuple
\begin{equation}
    r = (q, p, m, y, \hat{y}, l, s, n_{\text{in}}, n_{\text{out}}, c),
\end{equation}
where $q$ denotes the original query, $p$ the fully instantiated prompt (including system instructions and task-specific templates), $m$ the identifier of the candidate model, $y$ the raw model output, $\hat{y}$ the extracted answer used for scoring, $l$ the ground-truth label or reference answer, $s$ the corresponding evaluation score, $n_{\text{in}}$ and $n_{\text{out}}$ the input and output token counts, and $c$ the associated inference cost. In the performance-only setting, the cost component is omitted. Note that we keep $p$ identical across all candidate models for each dataset, so that comparisons focus on routing behavior rather than differences in prompt design.
\end{comment}

% \vspace{-1.3em}

\subsection{Modular Design}
To support flexible integration of diverse models, datasets, and routing algorithms, LLMRouterBench (see Fig.~\ref{fig:framework}) adopts a modular design built around three key modules: \textit{Collector}, \textit{Evaluator}, and \textit{Adaptor}. The Collector exposes a unified API to candidate LLMs, handling caching, retries, and cost tracking while consolidating all model outputs into the standardized format described above. The Evaluator implements rigorous, dataset-specific evaluation metrics (see Appendix Table~\ref{tab:benchmarks} for details) to ensure fair comparisons across models. The Adaptor converts the standardized format into algorithm-specific inputs for each routing method, maintaining consistent train-test splits given the same random seed. This modular design allows LLMRouterBench to interface with publicly available routing implementations with minimal modification, facilitating unified evaluation and faithful alignment with their original implementations.

\subsection{Evaluation Metrics}
\label{sec:metrics}

\paragraph{Performance metrics}
Let $\mathcal{D}$ be the set of evaluation datasets, with $d\in \mathcal{D}$ a generic dataset. For each routing method $a$, let  $\mathrm{Acc}(a,d)$ denote its accuracy on $d$, computed based on the correctness of the answers produced by the model selected by 
$a$ for each query in $d$. The Average Accuracy ($\mathrm{AvgAcc}$) for each routing method $a$ is then given by
$\mathrm{AvgAcc}
=
\frac{1}{|\mathcal{D}|}
\sum_{d \in \mathcal{D}} \mathrm{Acc}(a,d).$

We propose to compare each routing method to three baselines:
(1) {Random Router ($\mathcal{R}$)} randomly selects a model from the candidate pool per instance;
(2) {Best Single ($\mathcal{B}$)} selects a single model with the highest average accuracy across all datasets in hindsight;
(3) {Oracle ($\mathcal{O}$)} selects, for each instance, a model that yields a correct prediction if such a model exists in hindsight, choosing the one with minimum cost when multiple exist.
Random Router serves as a lower-bound baseline, while Best Single and Oracle provide meaningful reference points for achievable performance. We propose three metrics based on these baselines.
We define $\mathrm{Gain@\mathrm{b}} = \frac{1}{|\mathcal{D}|} \sum_{d \in \mathcal{D}} \left( \frac{\mathrm{Acc}(a,d)}{\mathrm{Acc}(\mathrm{b},d)} - 1 \right)$ and
$\mathrm{Gap@\mathcal{O}} = \frac{1}{|\mathcal{D}|} \sum_{d \in \mathcal{D}} \left( 1 - \frac{\mathrm{Acc}(a,d)}{\mathrm{Acc}(\mathcal{O},d)} \right)$,
where $\mathrm{b}\in\{\mathcal{R},\mathcal{B}\}$.
While $\mathrm{Gain@\mathcal{R}}$ and $\mathrm{Gain@\mathcal{B}}$ measure relative performance gains over Random Router and Best Single, respectively, $\mathrm{Gap@\mathcal{O}}$ measures the gap to the Oracle.

\paragraph{Performance-cost metrics} In the performance-cost setting, each routing method $a$ typically has a tunable parameter inducing configurations $\Theta$ with different performance-cost tradeoffs. For a configuration $\theta \in \Theta$, we denote its total inference cost by $\mathrm{Cost}(\theta)$. We compare each routing method to the Best Single $\mathcal{B}$, focusing on two configurations: (i) $\theta^\ast = \arg \max_{\theta \in \Theta}\mathrm{AvgAcc}(\theta)$, the configuration with the highest average accuracy (ignoring cost); and (ii) $\theta^\dagger = \arg\min_{\theta \in \Theta}\mathrm{Cost}(\theta)\text{ with }\mathrm{AvgAcc}(\theta)\ge\mathrm{AvgAcc}(\mathcal{B})$, the least-cost configuration with accuracy no worse than $\mathrm{AvgAcc}(\mathcal{B})$. Based on these, we define two metrics: $\mathrm{PerfGain}=\frac{\mathrm{AvgAcc}(\theta^\ast)}{\mathrm{AvgAcc}(\mathcal{B})}-1$, measuring the best achievable performance improvement, and $\mathrm{CostSave}=1-\frac{\mathrm{Cost}(\theta^\dagger)}{\mathrm{Cost}(\mathcal{B})}$, measuring the maximal cost reduction without sacrificing performance relative to the Best Single.

% \begin{equation}
%     \mathrm{PerfGain}
%     =
%     \frac{
%     \frac{1}{|\mathcal{D}|}\sum_{d\in\mathcal{D}}\mathrm{Acc}(\theta^\ast,d)
%     }{
%     \frac{1}{|\mathcal{D}|}\sum_{d\in\mathcal{D}}\mathrm{Acc}(\mathcal{B},d)
%     }
%     -1,
%     \label{eq:perf-gain}
% \end{equation}
% \begin{equation}
%     \mathrm{CostSave}
%     =
%     1-
%     \frac{
%     \frac{1}{|\mathcal{D}|}\sum_{d\in\mathcal{D}}\mathrm{Cost}(\theta^\dagger,d)
%     }{
%     \frac{1}{|\mathcal{D}|}\sum_{d\in\mathcal{D}}\mathrm{Cost}(\mathcal{B},d)
%     },
%     \label{eq:cost-save}
% \end{equation}

%To further compare performance-cost tradeoffs across methods using \textit{all} their available configurations, 
In addition, we perform a \textit{Pareto analysis}, a technique for evaluating tradeoffs in multi-objective optimization.
We define $S$ to include (i) all routing method configurations and (ii) all single models.
We say that $x\in S$ \textit{Pareto-dominates} $y \in S$ if
$\mathrm{AvgAcc}(x)\ge \mathrm{AvgAcc}(y)$ and $\mathrm{Cost}(x)\le \mathrm{Cost}(y)$,
and at least one inequality is strict.
The \textit{Pareto frontier} $\mathcal{P}$ is the set of configurations in $S$ that are not Pareto-dominated by any other configuration in $S$.
%Intuitively, this frontier contains all the non-inferior tradeoff configurations, each \emph{cannot} be simultaneously cheaper and more accurate than any other configuration.
For each routing method, we measure its  distance to the Pareto frontier. 
%We normalize $\mathrm{AvgAcc}$ and $\mathrm{Cost}$ over $S$ to the $[0,1]$ range, and 
Let $\tilde{\theta}$ and $\tilde{y}^*$ denote the normalized coordinates and frontier configurations, respectively. For a routing method $a$ with configurations $\Theta$, we define the distance as $\mathrm{ParetoDist} = \frac{1}{|\Theta|}\sum_{\theta \in \Theta}\min_{y^*\in\mathcal{P}}\|\tilde{\theta} - \tilde{y}^*\|_1$. Smaller values of $\mathrm{ParetoDist}$ indicate that the configurations of the method are, on average, closer to the Pareto frontier.

\begin{figure*}[t]
    \centering   
    \includegraphics[width=0.98\linewidth]{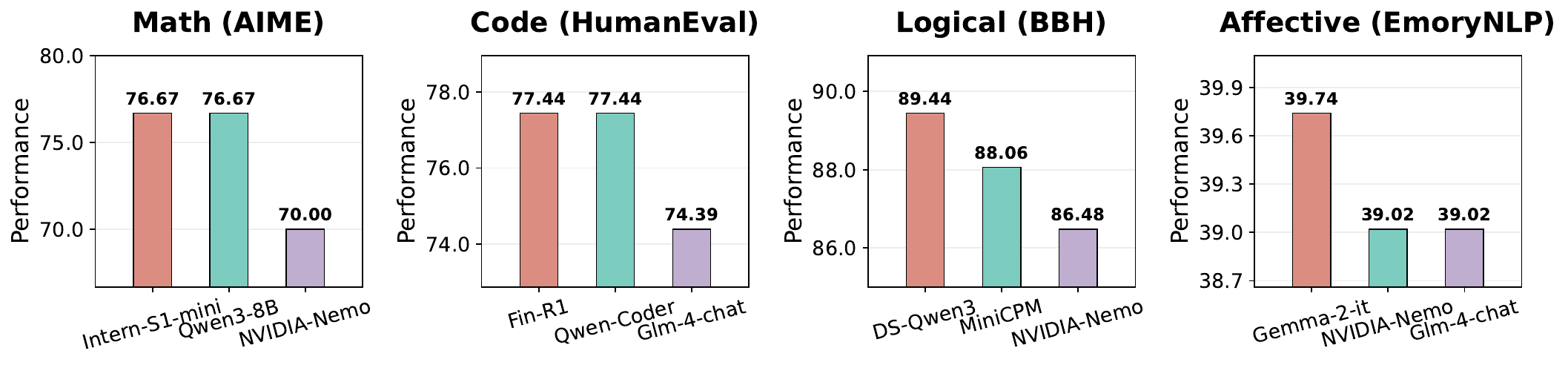}
    \vspace{-0.5em}
    \caption{Model performance across domains. No single model dominates all domains.
    }
    \label{fig:perf-main}
\end{figure*}

\subsection{Overall Benchmark Statistics}

Table~\ref{tab:dataset-overview} summarizes key statistics of LLMRouterBench across two routing settings.
In the performance-oriented setting, we evaluate 20 lightweight ($\sim$7B) models on 15 challenging datasets, totaling 745.0M tokens. The performance-cost setting assesses 13 flagship models across 10 datasets, totaling 1,030.3M tokens.
Taken together, LLMRouterBench comprises 23,945 prompts, 391,645 instances, and approximately 1.8B tokens, requiring substantial data collection efforts (1K GPU hours for lightweight inference and \$2,771.84 in API costs).
Notably, compared to recent studies, LLMRouterBench contains approximately \textbf{17$\times$} more total tokens (1.8B vs.\ 104M) than FusionFactory~\citep{feng2025fusionfactory}, and about \textbf{3$\times$} more prompts (23{,}945 vs.\ 8{,}400) than RouterArena~\citep{lu2025routerarena}. 
\begin{table}[h]
    \small
    \setlength{\tabcolsep}{6pt}
    \centering
    \begin{tabular}{@{}lcc@{}}
    \toprule
    \textbf{Metric} & \textbf{Performance-Oriented} & \makecell[c]{\textbf{Performance-Cost}} \\
    \midrule
    Datasets & 15 & 10 \\
    Models & 20 & 13 \\
    Prompts & 11{,}480 & 12{,}446 \\
    Instances & 229{,}600 & 161{,}520 \\
    % Prompt tokens (M) & 107.4 & 565.7 \\
    % Completion tokens (M) & 637.6 & 464.6 \\
    Tokens (M) & 745.0 & 1{,}030.3 \\
    Total cost & A800 1000 GPU hours & \$2{,}771.84$^\dagger$ \\
    \bottomrule
    \end{tabular}
    \caption{Overall statistics of LLMRouterBench. $^\dagger$About \$500 comes from LLM-based judging.
   %  Token counts are reported in millions (M); 
   %  costs reflect A800 80G GPU hours or API spending. 
   }
   \vspace{-2em}
    \label{tab:dataset-overview}
\end{table}

\section{Experiments}
\label{sec:experiments}
\subsection{Experimental Setup}
We evaluate 9 representative routing methods from recent published studies with available open-source implementations, and additionally include OpenRouter as a commercial router:
\begin{itemize}[leftmargin=*,itemsep=0.15em, topsep=0.2em]
    \item 
Performance-oriented setting:
RouterDC~\cite{chen2024routerdc}, EmbedLLM~\cite{zhuang2024embedllm}, MODEL-SAT~\cite{zhang2025capability}, GraphRouter~\cite{feng2025graphrouter}, and Avengers~\cite{zhang2025avengers}.
\item Performance-cost setting: HybridLLM~\cite{ding2024hybrid}, FrugalGPT~\cite{chen2024frugalgpt}, RouterLLM~\cite{ong2024routellm}, GraphRouter~\cite{feng2025graphrouter},  Avengers-Pro~\cite{zhang2025beyond}, and OpenRouter.
\end{itemize}
Appendix~\ref{app:experiment} and ~\ref{app:Baselines} provides detailed descriptions and implementation details.

\subsection{Results and Analyses}
Due to the lack of space, we present detailed benchmark results across all evaluated models and datasets in Appendix Tables~\ref{tab:benchmark-results}, \ref{tab:llm-benchmark-results-categorized}, and \ref{tab:llm-cost-comparison}. The reported results are averaged across runs; Appendix Tables~\ref{tab:benchmark-results-app}, \ref{tab:llm-benchmark-results-categorized-app}, and \ref{tab:llm-cost-app} additionally report per-dataset results averaged over five runs with different random seeds. In the following, we highlight and discuss the key findings.
\subsubsection{Performance-Oriented Setting}
\paragraph{No single model rules every domain; models exhibit complementary strengths.}
A central premise of performance-oriented routing is that different models excel in various domains. As shown in Fig.~\ref{fig:perf-main}, we find clear evidence of this complementarity: mathematics benchmarks are often led by models such as Intern-S1-mini or Qwen3-8B, code benchmarks by Qwen-Coder or Fin-R1, and logical and affective benchmarks by other models. This confirms the central premise of this field.

% \paragraph{Routers nearly match the \textit{Max Expert}.}
% Table~\ref{tab:perf-metrics-imp} reports accuracy and improvement metrics (Section~\ref{sec:metrics}). All routing methods significantly outperform Random Router, achieving high $\mathrm{ImpRand}$ values. The \textit{Max Expert} baseline, selecting the best single model per dataset from Table~\ref{tab:benchmark-results}, achieves $73.10$ accuracy. All learned routers except RouterDC ($61.33$) approach this upper bound: EmbedLLM, GraphRouter, MODEL-SAT, and Avengers obtain $71.24$-$71.94$ accuracy, recovering about $96\%$-$98\%$ of \textit{Max Expert} with small-magnitude $\mathrm{ImpMax}$ values. Notably, the training-free, clustering-based Avengers reaches $71.94$ accuracy, only $1.16$ points below \textit{Max Expert}. % For trainable routers (marked with $\dagger$), we report peak test performance.

%\paragraph{With routing, smaller models can collectively outperform much larger ones.} Although the best single 7B model (xxx; highest AvgAcc) still trails a much larger model by a clear margin, routing changes the picture: a routed pool of 7B models can collectively surpass the larger model, even though none of the individual models can do so alone.

\begin{figure*}[t]
    \centering   
    \includegraphics[width=\linewidth]{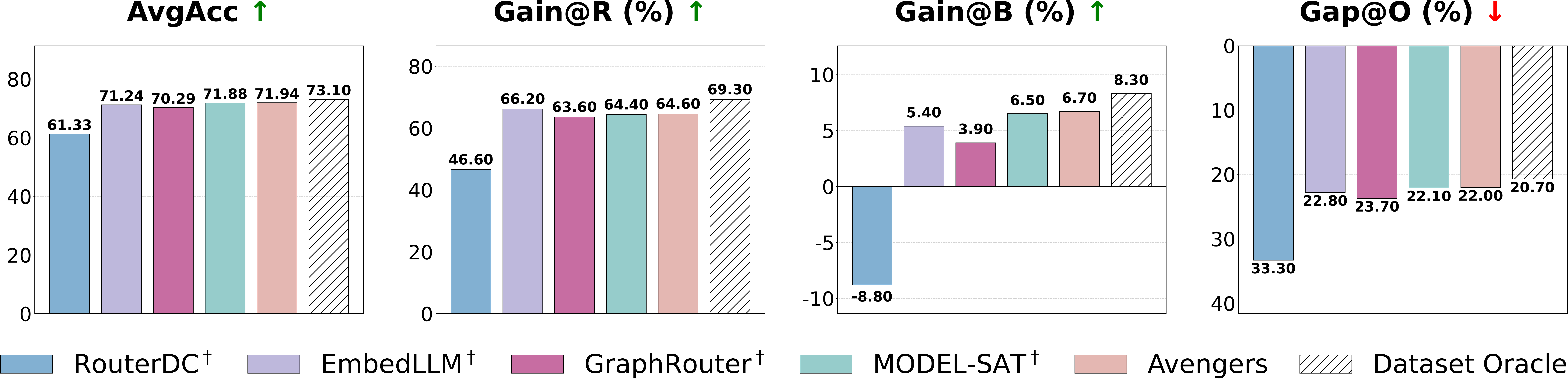}
    \vspace{-1.6em}
    \caption{Performance metrics on LLMRouterBench. 
    $^\dagger$Peak score on the test set. 
    }
    \vspace{-1.em}
    \label{fig:perf-metrics-viz}
\end{figure*}

\begin{figure}[t]
    \centering   
    \includegraphics[width=\linewidth]{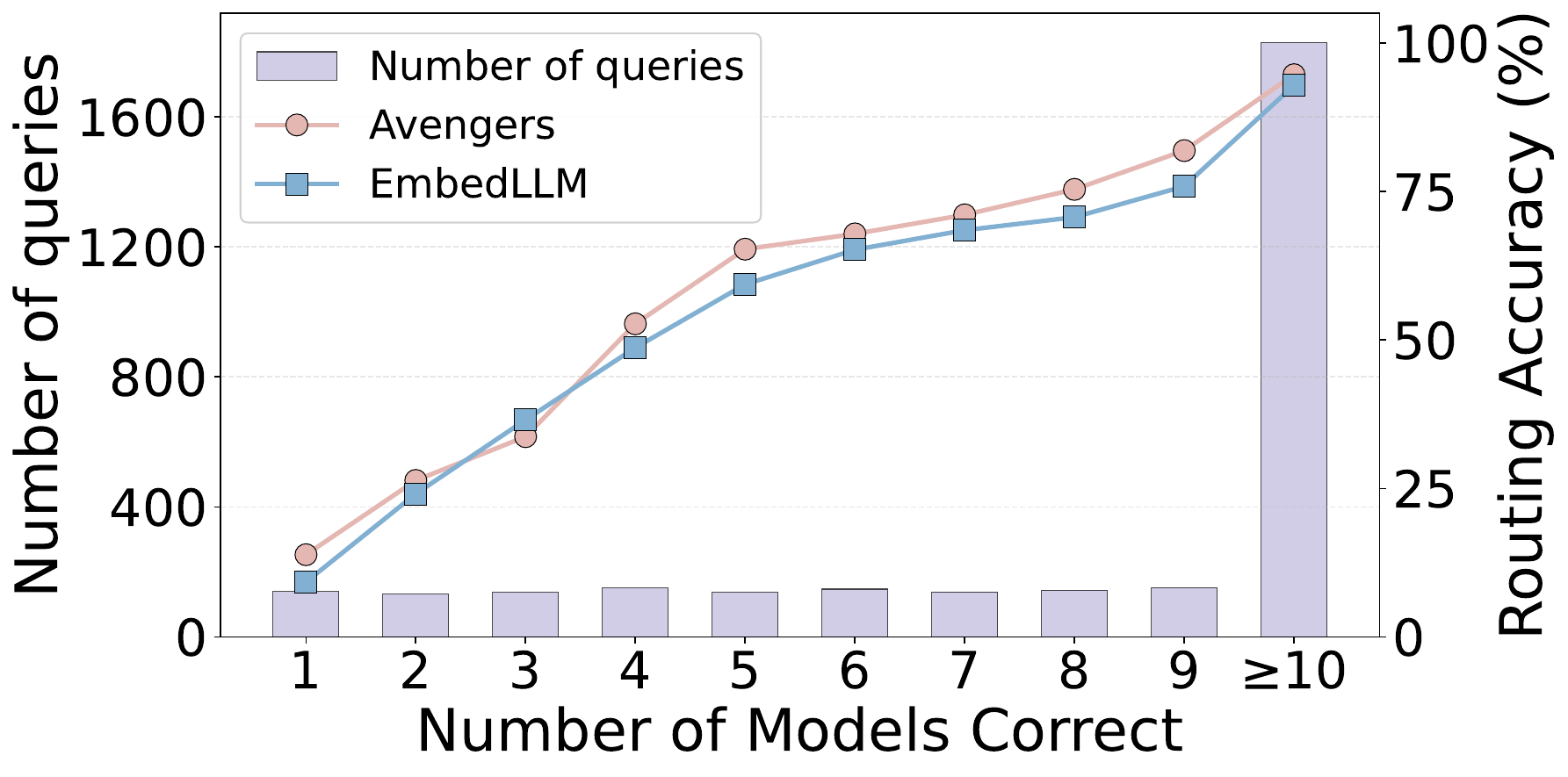}
    \caption{Query hardness distribution (by number of correct models) with router accuracy.
    }
    \vspace{-2em}
    \label{fig:oracle-anslysis}
\end{figure}

\paragraph{Top routing methods are comparable, but can be free of neural network training.} 
LLM routing is not new, resulting in a substantial body of literature. Nonetheless, our results indicate that, despite continued methodological innovation, many routing approaches achieve broadly comparable performance in practice. As shown in Fig.~\ref{fig:perf-metrics-viz}, across multiple metrics ($\mathrm{Gain@\mathcal{R}}$, $\mathrm{AvgAcc}$, $\mathrm{Gain@\mathcal{B}}$, and $\mathrm{Gap@\mathcal{O}}$), leading routing methods (EmbedLLM, GraphRouter, MODEL-SAT, and Avengers) yield similar outcomes. Note that the Avengers achieve such performance primarily through clustering and do not require neural-network training.

This observation has two implications. First, it is favorable from a deployment standpoint: lightweight routers, which are inexpensive to develop and straightforward to update, may be sufficient in practice. On the other hand, the lack of differentiation among leading methods suggests that a large fraction of routing gains may be attributable to capturing coarse-grained domain structure (e.g., distinguishing mathematics and code) rather than learning highly nuanced decision boundaries. This is supported by the proximity of these methods to the Dataset Oracle (the hatched bar in Fig.~\ref{fig:perf-metrics-viz}), which assigns each dataset to the single model with the highest accuracy on that dataset.

\paragraph{Routing remains far from its capability ceiling: current methods often miss the lone correct model.} Although top routing methods perform similarly, which may raise the question of whether further gains are available, The GaptoOracle, shown in Fig.~\ref{fig:perf-metrics-viz} (d), reveals a significant gap to the routing upper bound given by the Oracle baseline.
A key contributor is model-recall failure: when only one or a few candidate models produce the correct answer, current routers often fail to select them. As analyzed in Fig.~\ref{fig:oracle-anslysis}, this issue accounts for a non-trivial portion of the remaining error. For example, on queries where at most three experts answer correctly (410 queries, 11.9\% of the test set), Avengers and EmbedLLM achieve low accuracy (24.6\% and 23.2\%, respectively).

Closing this gap will likely require routers that more reliably detect and prioritize these cases, for example via improved uncertainty or difficulty estimation, or explicit mechanisms to boost recall of rare-but-critical experts, which
constitutes a promising direction for future work.
%.

\paragraph{Embedding models have little influence on routing performance.} Because many routing methods rely on an embedding model, we examine how this choice affects performance. Surprisingly, the embedding models have consistently little differences across routing methods. Appendix Table~\ref{tab:embedding-model-comparison} shows that by replacing \textit{gte-qwen2-7B-instruct} with \textit{nli-bert-base}~\cite{reimers-2019-sentence-bert} and \textit{all-MiniLM-L6-v2}~\cite{wang2020minilm}, we do not observe a significant difference in performance among GraphRouter, EmbedLLM, and Avengers, all of which depend on embeddings. Note that both alternatives are weaker backbones: \textit{all-MiniLM-L6-v2} involves only 22.7M parameters, and \textit{nli-bert-base} is deprecated in Sentence-Transformers for poor sentence-embedding quality.

This suggests that embedding quality may not be the primary bottleneck for current embedding-based routers. Future gains may come less from improving semantic representations per se, but more from developing routing mechanisms that better translate representations into reliable model selection, particularly under distribution shift and in rare cases where correct performance hinges on selecting a specific specialist model.

%more from better decision rules and reliability calibration.
%In fact, the effect of the embedding choice sometimes may exceed the gains from changing the router design itself. This points to a direction for future work: stronger embedding models particularly for routing, potentially ones that explicitly encode instance difficulty.

\begin{figure}[t]
    \centering   
    \includegraphics[width=\linewidth]{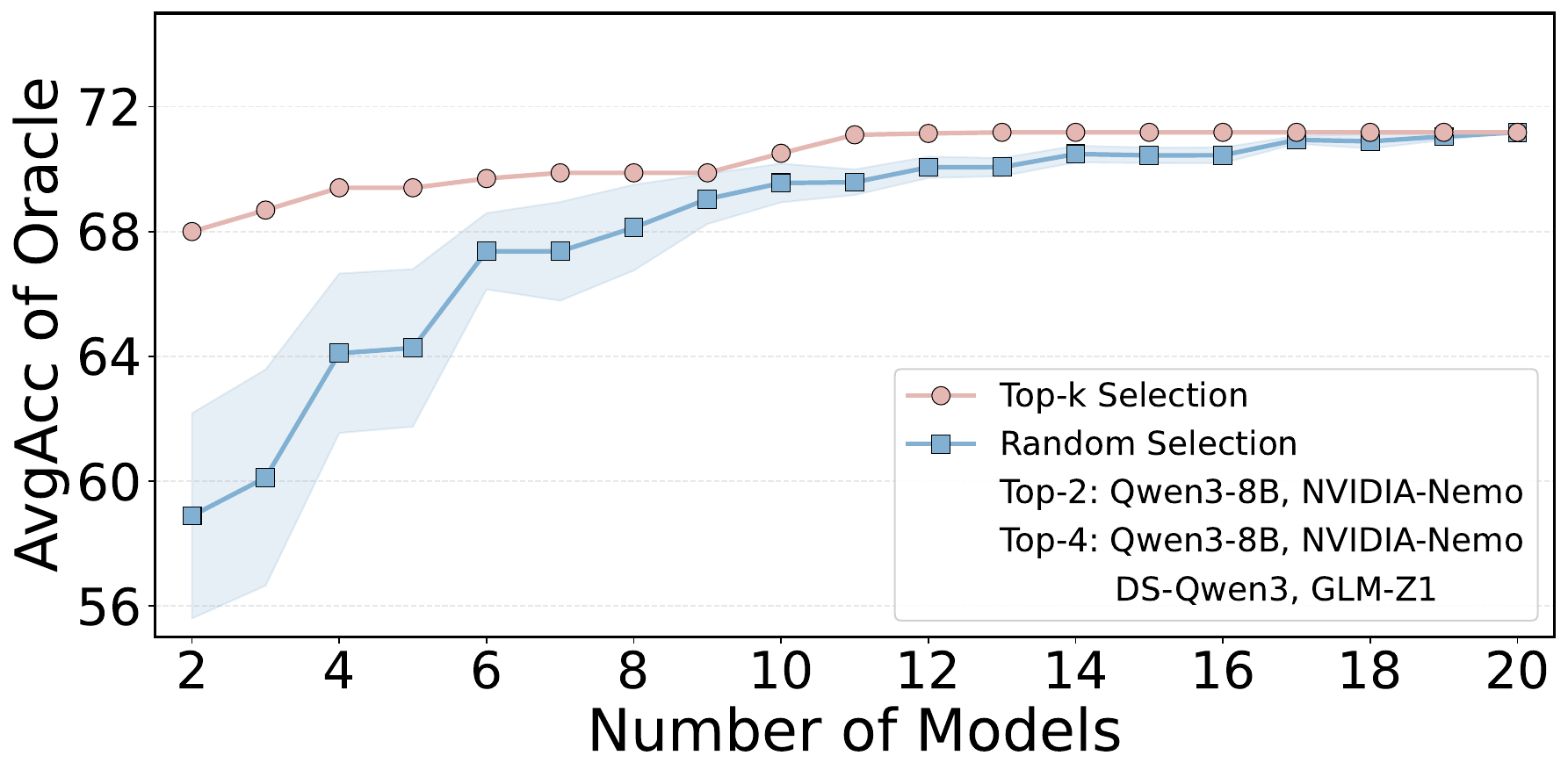}
    \vspace{-1.5em}
    \caption{Comparison of Oracle performance under selected versus random subsets. 
    Adding more models yields diminishing returns, but a well-chosen small subset can outperform a larger random pool.
    }
    \label{fig:TopK_vs_RandomK}
\vspace{-1.5em}
\end{figure}

\paragraph{Adding more models yields diminishing returns, but a well-chosen subset can make a difference.} A common hypothesis in the field~\cite{jiang-etal-2023-llm} is that expanding the candidate pool should increase complementarity and therefore improve collective performance. However, using the Oracle baseline, we find clear diminishing returns as more models are added (Fig.~\ref{fig:TopK_vs_RandomK}). The largest gains occur when moving from a very small pool to a moderate one, after which additional models contribute only marginal improvements. By contrast, how we select a small subset matters substantially: comparing random selection to choosing the top-k models by average accuracy, the latter consistently delivers stronger performance.

This suggests that careful curation can outweigh simply scaling the pool; 
 a carefully selected moderate pool plus a robust router may offer most of the achievable benefit without the overhead of maintaining a very large pool.
Future work should study model pool curation jointly with routing: selecting a small set that maximizes complementarity, potentially via coverage or diversity-style selection.

\subsubsection{Performance-Cost Settings} 
\paragraph{Models complement each other on performance and cost-efficiency.}
Similar to the performance-oriented setting, we observe complementarity among models in terms of performance and cost. As Appendix Table~\ref{tab:llm-benchmark-results-categorized} shows, while GPT-5 achieves the best average accuracy and leads on HLE, substantially cheaper models such as Qwen3-235B and DeepSeek-R1 can match proprietary-level accuracy on selected mathematics and QA tasks. These results support the central premise behind routing for performance-cost tradeoffs.
\begin{figure}[t]
    \centering   
    \includegraphics[width=0.95\linewidth]{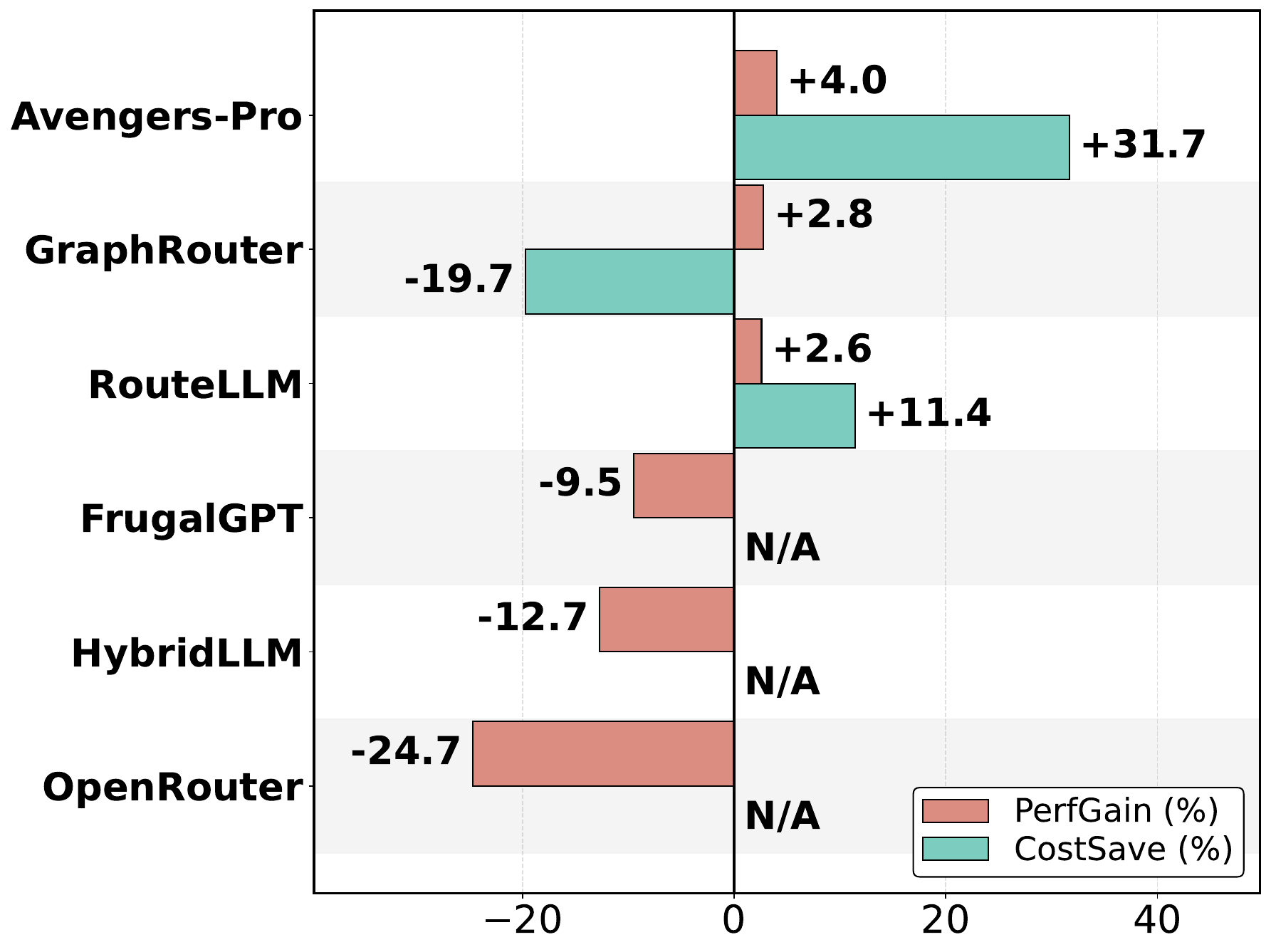}
    \vspace{-0.5em}
    \caption{ Performance gains and cost savings of various routing methods relative to the GPT-5. Cost savings are reported only for methods achieving accuracy equal to or higher than GPT-5; otherwise marked as N/A.}
    \label{fig:perf-cost-1}
    \vspace{-1.3em}
\end{figure}

\paragraph{Effective routing improves upon the Best Single and reduces cost without sacrificing performance, but not all routers succeed.}
Leveraging model complementarity, as shown in  Fig.~\ref{fig:perf-cost-1}, top routing methods achieve up to a 4\% average accuracy gain over the Best Single model and up to a 31.7\% cost reduction while matching Best Single performance. However, these gains are not universal: several routers fail to outperform the Best Single, and some (especially binary routers, such as HybridLLM and FrugalGPT) struggle to trade cost for savings while preserving Best Single accuracy. Notably,  the commercial router OpenRouter yields the smallest (indeed, negative) performance improvement (-24.7\%) relative to the Best Single model and fails to match its performance, despite operating over a much larger candidate pool.\footnote{OpenRouter uses a platform-defined model pool that differs from ours and is not user-configurable~\cite{openrouter-auto}; we provide details for their pool in Appendix Table~\ref{tab:openrouter-model-pool}.} 

These results reinforce the appropriateness of Best Single as a baseline: a non-trivial fraction of routing strategies do not outperform this simple alternative. More broadly, the variability across routers indicates substantial remaining headroom for robust performance-cost routing, particularly for methods that can deliver cost reductions without sacrificing accuracy.

\vspace{-0.2em}

\begin{figure}[t]
    \centering
    \hspace{-0.8cm}  % 向左移动0.5厘米，可根据需要调整数值
    \includegraphics[width=0.98\linewidth]{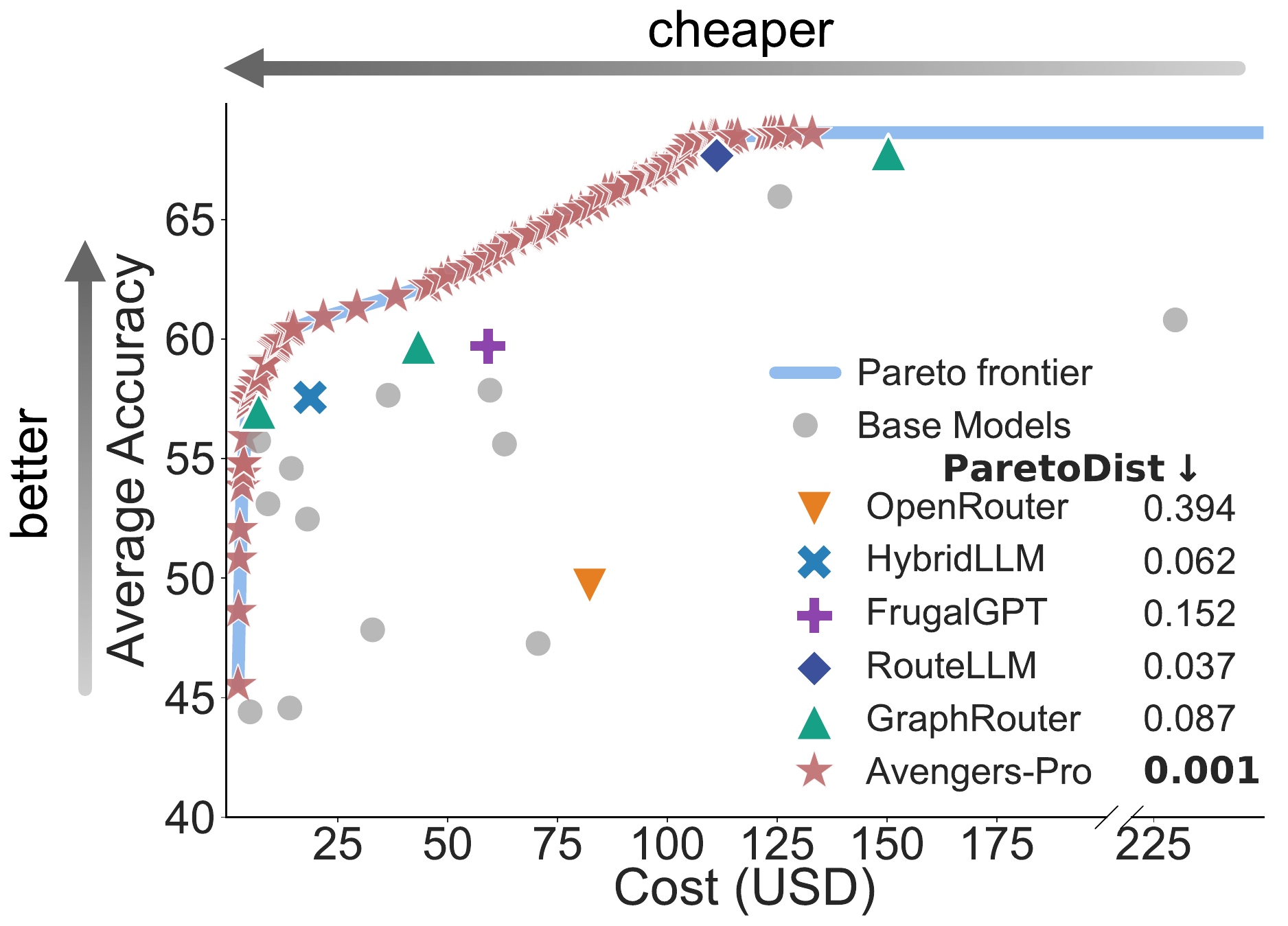}
        \vspace{-0.6em}
    \caption{Average accuracy versus total inference cost for all base models and routing methods, with the empirical Pareto frontier highlighted.}
    \vspace{-1.3em}
    \label{fig:Pareto}
\end{figure}

\paragraph{Avengers-Pro achieves a Pareto-optimal performance-cost tradeoff.}
We visualize the Pareto frontier, a stylish analysis for multi-objective optimization in Fig.~\ref{fig:Pareto}. Unlike the performance-oriented setting (where several routing methods are competitive), Avengers-Pro nearly dominates the frontier. That is, relative to any single model or other routing methods, Avengers-Pro is almost always cheaper at comparable performance or achieves higher performance at comparable cost. No other methods or single models can be simultaneously cheaper and more accurate than Avengers-Pro.
This is further supported by the ParetoDist metric in Fig.~\ref{fig:Pareto}; Avengers-Pro attains ParetoDist near zero, generally being Pareto-optimal, whereas other routers exhibit substantially larger distances.

These results have two implications. First, they demonstrate that strong performance-cost routing is achievable in practice: the Pareto frontier is not merely a theoretical construct, and a well-designed router can operate near it. Second, they indicate that future progress should be assessed not only by average accuracy, but also by whether new methods shift the frontier—that is, expand the set of attainable operating points toward simultaneously lower cost and higher performance.

\vspace{-0.3em}

\paragraph{LLMRouterBench enables extending routing to performance-cost-latency optimization.}
We note that LLMRouterBench also enables latency-aware analysis.
By tracking the number of consumed tokens (supported by our LLMRouterBench) and combining this with standard serving statistics (provided by OpenRouter, e.g., time-to-first-token and tokens-per-second), one can approximate end-to-end response time for different models under a unified protocol. To date, however, routing research has largely focused on optimizing accuracy and/or cost, and, to our knowledge, we do not notice any methods that explicitly target the joint performance-cost-latency tradeoffs.
\begin{figure}[t]
    \centering   
    \includegraphics[width=\linewidth]{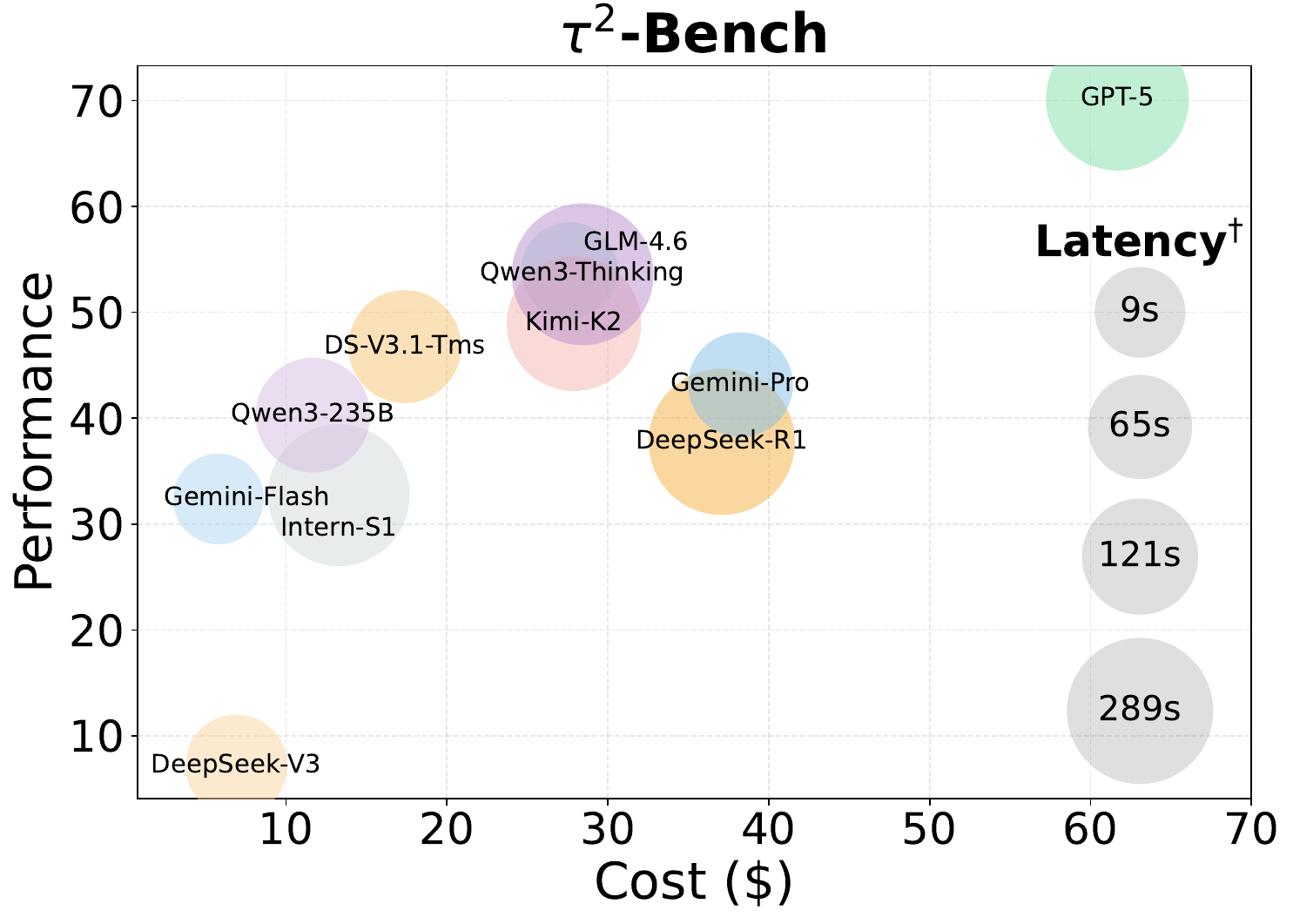}
    \vspace{-2em}
    \caption{Model performance-cost-latency tradeoffs on $\tau^2$-Bench. $^\dagger$ Latency measurements were collected from the OpenRouter platform (Jan. 5, 2026).}
    \label{fig:openrouter-latency}
    \vspace{-1.2em}
\end{figure}

In Fig.~\ref{fig:openrouter-latency}, we illustrate how different models vary along these three dimensions. Models that are similar in accuracy and cost can nonetheless differ markedly in latency. For example, Qwen3-Thinking and GLM-4.6 have similar performance and cost but differ notably in latency (262.1s vs. 32.4s).  This introduces an additional axis of complementarity that is directly relevant to user experience. From a practical standpoint, when two models deliver comparable accuracy at similar cost, users will often prefer the model that responds faster. Accordingly, our dataset provides the necessary signals to support systematic exploration of this tri-objective setting.

\section{Conclusions}

We present LLMRouterBench, a large-scale benchmark and unified framework for LLM routing under both performance-oriented and performance-cost trade-off settings. By consolidating over 400K instances spanning 21 datasets and 33 models, and by introducing comprehensive metrics together with 10 representative baselines, LLMRouterBench provides a solid foundation for systematic study in this rapidly evolving area.
Our results reaffirm the central motivation of LLM routing—strong complementarity across models—but also challenge several prevailing claims in the literature. Under a unified evaluation, most routing methods collapse to similar performance, and multiple recent approaches, including widely deployed commercial routers, fail to reliably outperform a simple baseline. A large and persistent gap to the Oracle remains, driven primarily by systematic model-recall failures rather than insufficient ensemble capacity. We further demonstrate that common design choices, such as the selection of backbone embedding models or aggressive scaling of ensemble size, yield limited gains in practice. Additionally, we show that our benchmark enables extending routing to tradeoffs for latency. 

%We hope LLMRouterBench will serve as a common ground for fair comparison, deeper diagnosis, and more accessible and reproducible research, ultimately accelerating progress toward reliable, efficient, and deployable LLM routing systems.

\section*{Limitations}
\label{sec:limitations}
Our work has three limitations. First, while we evaluate a broad set of routing methods, we do not cover all existing approaches. Given the large and growing number of routing methods, we focus on recent approaches with publicly available implementations. LLMRouterBench is modular by design, allowing additional routers to be integrated via lightweight adapters without reimplementing the full pipeline.

Second, the benchmark is built from 21 widely used datasets that reflect common evaluation regimes for contemporary lightweight and flagship LLMs. Other settings, such as domain-specific verticals, very long-context tasks, and multimodal benchmarks, are not included. The routing formulations, metrics, and analysis procedures used in LLMRouterBench are generic and can be applied to these settings by adding new dataset evaluators.

Third, the latency analysis is approximate. We estimate latency using token-level usage statistics together with throughput figures reported by OpenRouter. These estimates correspond to a specific provider configuration and should be interpreted as indicative rather than definitive.

\section*{Acknowledgements}
This work was supported by the Shanghai Municipal Science and Technology Major Project.

% Bibliography entries for the entire Anthology, followed by custom entries
%\bibliography{anthology,custom}
% Custom bibliography entries only
\bibliography{custom}

% \clearpage
\appendix
\section{Further Related Work}
\paragraph{Routing Benchmark.}
% Recent studies have attempted to establish benchmarks for evaluating routing methods. 
% RouterBench~\cite{hu2024routerbench} provides a benchmark for evaluating multi-LLM routing methods, containing data from early-generation LLMs and 8 simple datasets.
% EmbedLLM~\cite{zhuang2024embedllm} collects routing data from open-source models, focusing solely on performance without incorporating proprietary models or cost information.
% RouterEval~\cite{huang2025routereval} evaluates a large number of open-source models across 12 datasets to benchmark routing methods, without incorporating cost-related analysis.
% FusionFactory~\cite{feng2025fusionfactory} collects routing data from 20 open-source LLMs evaluated on 14 public datasets, with inference cost estimated based on Together AI pricing.
% RouterArena~\cite{lu2025routerarena} constructs a dataset to evaluate different LLM routing systems, but uses inconsistent model pools across routers, leading to incomparable results and limiting the validity for fair evaluation.
% Existing benchmarks (Tables~\ref{tab:router-benchmarks} and~\ref{tab:routing-benchmarks}) remain limited to a few easy datasets already saturated by leading models, focus mainly on open-source LLMs without proprietary or realistic cost data, and lack unified interfaces for fair routing evaluation.

Recent work has proposed several benchmarks for evaluating LLM routing. RouterBench~\cite{hu2024routerbench} targets multi-LLM routing but is restricted to early-generation models and eight relatively simple datasets. 
% EmbedLLM~\cite{zhuang2024embedllm}, RouterEval~\cite{huang2025routereval}, and FusionFactory~\cite{feng2025fusionfactory} mainly benchmark routing over open-source models: EmbedLLM and RouterEval ignore inference cost altogether, whereas FusionFactory approximates it using provider-specific prices (e.g., Together AI). 
EmbedLLM~\cite{zhuang2024embedllm}, RouterEval~\cite{huang2025routereval}, and FusionFactory~\cite{feng2025fusionfactory} benchmark routing over open-source models, with EmbedLLM and RouterEval providing no inference cost information.
RouterArena treats routing systems as black boxes and constructs a dataset for comparing routing systems. However, it uses different model pools across routers, undermining cross-method comparability, and does not provide per-prompt, per-model data. 
As summarized in Tables~\ref{tab:router-benchmarks} and Appendix Table~\ref{tab:routing-benchmarks}, existing benchmarks narrowly focus on a small number of relatively easy tasks, lack coverage of flagship models with realistic inference costs, and do not provide a unified interface for fair comparisons across routing methods.
% These gaps motivate \textsc{LLMRouterBench}, a massive benchmark spanning diverse and challenging tasks, jointly evaluating open-source and proprietary models with realistic costs, and providing a unified routing evaluation interface.
These gaps motivate LLMRouterBench, a massive routing benchmark that spans diverse and challenging tasks, jointly evaluates lightweight and flagship models under realistic costs, and offers a unified interface for plug-and-play comparison of routing methods.
\section{Implementation Details}
\label{sec:appendix}
\subsection{Data Collection}
All $\sim$7B open-source models are deployed on NVIDIA A800-80G GPUs using vLLM~0.8.4 for efficient batched inference. Flagship model outputs are collected via OpenRouter, with the exception of GLM-4.6 and Intern-S1, which are accessed through official APIs. All model generations use temperature 0.2 and top\_p 1.0, with remaining decoding parameters set to their default values. Each API request is retried up to 10 times upon failure; requests exceeding this limit are marked as failures and assigned a score of 0. 

\subsection{Experimental Setup}
\label{app:experiment}
To ensure fair and consistent comparisons across baselines, we standardize the experimental setup as follows:
(i) All experiments are conducted using a 70\% training and 30\% test split, repeated five times with different random seeds (42, 999, 2024, 2025, and 3407). 
% (2) Trainable router methods report peak test performance rather than using fixed training steps.
(ii) Embedding-dependent methods uniformly utilize \textit{gte-qwen2-7B-instruct}~\cite{li2023gteqwen} embeddings.
(iii) For binary routers (RouteLLM, HybridLLM, FrugalGPT), we route between Qwen3-235B and GPT-5, as GPT-5 is the strongest model in our pool while Qwen3-235B offers competitive accuracy at much lower cost.
All datasets and models used in this paper are publicly available and properly cited. Our usage complies with their original licenses and intended research purposes.

\subsection{Baselines}
\label{app:Baselines}
\paragraph{RouterDC}
We adopt the official implementation of RouterDC~\citep{chen2024routerdc}, replacing the original encoder with \texttt{gte-qwen2-7B-instruct} to enable fair comparison under a unified embedding backbone. Training is conducted using DeepSpeed with distributed multi-GPU parallelism across eight NVIDIA A800-80G GPUs, with a per-device batch size of 8. All other hyperparameters remain consistent with the original configuration. The resulting model contains approximately 7B trainable parameters. % The accuracy curves for both training and test are shown in Figure \ref{fig:routerdc_acc}.

\paragraph{MODEL-SAT} % ~\citep{zhang2025capability}
Due to the incomplete release of the official implementation~\citep{zhang2025capability}, we re-implement the core components of MODEL-SAT. We use \texttt{gte-qwen2-7B-instruct} as the embedding model and \texttt{Qwen2.5-7B-Instruct} as the language model, connected via a two-layer MLP projector. Training is performed using DeepSpeed with data parallelism over eight NVIDIA A800-80G GPUs, using a per-device batch size of 4. The learning rate is set to 1e-6 for both the embedding and language models, and 1e-5 for the projector. We first fine-tune only the projector for approximately 1{,}000 steps, then jointly fine-tune all components. A warmup ratio of 0.1 is used throughout. The final model contains roughly 14B trainable parameters. % Figure \ref{fig:model_sat_acc} illustrates the training and test accuracy curves.

\paragraph{EmbedLLM}
We use the official implementation of EmbedLLM~\citep{zhuang2024embedllm}, with query embeddings generated by \texttt{gte-qwen2-7B-instruct} to ensure consistency across methods. Input layer dimensions are adjusted accordingly. Training is conducted with an increased batch size of 32{,}768 for improved stability, while all other hyperparameters follow the original setting. The model has approximately 12M trainable parameters. % The training and test accuracy curves are presented in Figure \ref{fig:embedllm_acc}.

\paragraph{HybridLLM}
We adopt the official implementation of HybridLLM~\citep{ding2024hybrid}, substituting the original encoder with \texttt{gte-qwen2-7B-instruct} for consistency with other baselines. Training is performed using DeepSpeed with a distributed multi-GPU setup across eight NVIDIA A800-80G GPUs, with a batch size of 8 per device. All remaining hyperparameters follow the original configuration. We use \texttt{GPT-5} as the strong model and \texttt{Qwen3-235B} as the weak model throughout.

\paragraph{RouteLLM}
We adopt the official implementation of RouteLLM~\citep{ong2024routellm}, using the \emph{Matrix Factorization} (MF) router and \texttt{gte-qwen2-7B-instruct} to generate query embeddings. Throughout training and evaluation, we consistently use \texttt{GPT-5} as the strong model and \texttt{Qwen3-235B} as the weak model. Following the default setting in the official code, we set the win-rate threshold to \textbf{0.5}—that is, when the estimated win rate exceeds this threshold, we select the strong model; otherwise, we use the weak model. Under our configuration (\texttt{dim} = 128, projected from 3584-dimensional \texttt{gte-qwen2-7B-instruct} embeddings), the MF router itself is extremely lightweight, with only $\sim 4.6\times 10^5$ trainable parameters, and is trained for 100 epochs, making the routing overhead negligible compared to querying the underlying LLMs.

\paragraph{FrugalGPT}
We follow the official FrugalGPT training pipeline and fine-tune an embedding-based scorer derived from \texttt{gte-Qwen2-7B-Instruct}. Training is conducted for two epochs using AdamW with a linear warm-up schedule. We set the learning rate to \(3 \times 10^{-5}\), apply a weight decay of 0.01, fix the warm-up ratio at 0.03, and use a per-device batch size of 4. For routing, we adopt the FrugalGPT cascade strategy. Models are ranked by their average cost on the training set, and cascade evaluation is enabled with a default threshold of 0.5. Model-specific decision thresholds are learned automatically, and the cascade depth is capped at 2 models per query to balance cost and performance.
We use \texttt{GPT-5} as the strong model and \texttt{Qwen3-235B} as the weak model throughout.

\paragraph{GraphRouter}
We use the official implementation of GraphRouter~\citep{feng2025graphrouter}. All query, task, and model description embeddings are generated using \texttt{gte-qwen2-7B-instruct}. Input layer dimensions are adjusted accordingly. To mitigate label bias caused by tied predictions, we replace argmax-based one-hot labels with multi-hot supervision that includes all tie-optimal models per query. For improved training stability, we increase the number of training epochs to 10{,}000. The model has approximately 0.1M trainable parameters. We follow the original paper in employing its three configurations: \textit{Performance First (PF)}, \textit{Balance (BL)}, and \textit{Cost First (CF)}.

\paragraph{Avengers(-Pro)}
We adopt the official implementation of the clustering-based method proposed by~\citet{zhang2025beyond}. Query embeddings are generated using \texttt{gte-qwen2-7B-instruct}, and $k$-means clustering is applied with $k=64$. This method involves no neural network training. For Avengers-Pro, following the original paper, we vary the performance coefficient (i.e., $1{-}$cost coefficient) from 0 to 1 in increments of 0.01, resulting in 101 configurations.

\paragraph{OpenRouter}
We use the official API provided by OpenRouter with the \texttt{openroute/auto} model, which is currently an advanced commercial LLM router available. All model generations use temperature 0.2 and top\_p 1.0, with remaining decoding parameters set to their default values. Each API request is retried up to 10 times upon failure; requests exceeding this limit are marked as failures and assigned a score of 0. The detailed model pool supported by OpenRouter is summarized in Table~\ref{tab:openrouter-model-pool}. Note that GPT-5 and Gemini-2.5-pro, the strongest models used in LLMRouterBench, are included, and OpenRouter's model pool is even larger. We therefore treat OpenRouter as a representative commercial router baseline.

% Appendix: Benchmarks and their routing baselines
\begin{table*}[t]
\centering
\small
\setlength{\tabcolsep}{8pt}
\begin{tabular}{lc}
\toprule
\textbf{Benchmark} & \textbf{Routing Baseline} \\
\midrule
RouterBench & KnnRouter, MLPRouter, SvmRouter \\
EmbedLLM & EmbedLLM~\cite{zhuang2024embedllm} \\
RouterEval & KnnRouter, MLPRouter, Multi-class classification Router, RoBERTa-Kmeans Router\\
FusionFactory & KnnRouter, BertRouter, MLPRouter, SvmRouter, GraphRouter~\cite{feng2025graphrouter} \\
RouterArena & \xmark \\
\midrule
 LLMRouterBench    & \makecell[l]{%
  RouterDC~\cite{chen2024routerdc}, EmbedLLM~\cite{zhuang2024embedllm}, 
  Model-SAT~\cite{zhang2025capability}\\ Avengers~\cite{zhang2025avengers}, HybridLLM~\cite{ding2024hybrid}, FrugalGPT~\cite{chen2024frugalgpt}\\RouteLLM~\cite{ong2024routellm}, GraphRouter~\cite{feng2025graphrouter}, Avengers-Pro~\cite{zhang2025beyond} \\ OpenRouter~\cite{openrouter}
}  \\
\bottomrule
\end{tabular}
\caption{Benchmarks and the corresponding routing baselines provided in their official codebases.}
\label{tab:routing-benchmarks}
\end{table*}

% \begin{figure*}[t]
%     \centering    \includegraphics[width=\linewidth]{figs/figure3-perf_cost_sample.pdf}
%     \caption{Cost-performance tradeoffs on LLMRouterBench. (A) Instance-level accuracy versus total inference cost for all base models and routing methods, with the empirical Pareto frontier highlighted. (B) Performance gains and cost savings of various routing methods relative to the GPT-5 at instance-level. Cost savings are reported only for methods achieving accuracy equal to or higher than GPT-5; otherwise marked as N/A.}
%     \label{fig:perf-cost-sample}
% \end{figure*}

\begin{table*}[h]
\setlength{\tabcolsep}{1pt}
\small
\centering
\scriptsize
\resizebox{\textwidth}{!}{%
\begin{tabular}{lccccccccccccccc}
\toprule
\multirow{2}{*}{\textbf{Model}} &
\multicolumn{3}{c}{\textbf{Mathematics}} &
\multicolumn{3}{c}{\textbf{Code}} &
\multicolumn{3}{c}{\textbf{Logical}} &
\multicolumn{4}{c}{\textbf{Knowledge}} &
\multicolumn{2}{c}{\textbf{Affective}} \\
\cmidrule(lr){2-4}
\cmidrule(lr){5-7}
\cmidrule(lr){8-10}
\cmidrule(lr){11-14}
\cmidrule(lr){15-16}
&
\textbf{AIME} &
\textbf{M500.} &
\textbf{MBen.} &
\textbf{MBPP} &
\textbf{HE.} &
\textbf{LCB.} &
\textbf{KOR.} &
\textbf{K\&K.} &
\textbf{BBH} &
\textbf{MP.} &
\textbf{GPQA} &
\textbf{FQA.} &
\textbf{MQA.} &
\textbf{Emory.} &
\textbf{MELD} \\
\midrule
\textbf{DH-Llama3-it} & 0.00 & 29.80 & 38.67 & 54.00 & 51.22 & 13.08 & 31.92 & 15.57 & 49.07 & 36.66 & 27.78 & 49.96 & 57.58 & 38.16 & 47.16 \\
\textbf{DS-Qwen3} & 68.33 & \cellcolor{lightred}94.20 & 90.00 & 35.01 & 25.00 & 64.74 & 56.00 & 60.86 & \cellcolor{deepred}89.44 & \cellcolor{deepred}70.83 & \cellcolor{deepred}60.61 & 70.36 & \cellcolor{deepred}81.62 & 38.74 & 51.06 \\
\textbf{DS-Qwen} & 40.00 & 88.40 & 88.00 & 53.59 & 45.12 & 41.14 & 45.36 & 50.86 & 69.17 & 47.75 & 35.86 & 63.30 & 34.80 & 27.55 & 34.90 \\
\textbf{Fin-R1} & 11.67 & 75.40 & 66.67 & \cellcolor{lightred}68.99 & \cellcolor{deepred}77.44 & 6.82 & 33.20 & 19.00 & 61.57 & 48.65 & 27.27 & 68.70 & 63.24 & 34.72 & 50.41 \\
\textbf{GLM-Z1} & 61.67 & \cellcolor{deepred}94.60 & \cellcolor{deepred}95.33 & 62.94 & 60.37 & 61.52 & 53.84 & 47.57 & 84.26 & 67.33 & 56.06 & 68.70 & 71.01 & 35.01 & 45.86 \\
\textbf{Intern-S1-mini} & \cellcolor{deepred}76.67 & 91.40 & 64.67 & 44.05 & 43.29 & 26.26 & \cellcolor{deepred}58.16 & \cellcolor{deepred}79.43 & 83.06 & 56.04 & 48.99 & 70.36 & 68.42 & 36.87 & 50.41 \\
\textbf{Llama-3.1-it} & 8.33 & 49.80 & 46.67 & 60.68 & 61.59 & 16.78 & 24.16 & 13.71 & 58.70 & 46.65 & 26.26 & 52.92 & 68.26 & 33.72 & 48.05 \\
\textbf{UltraMedical} & 0.00 & 42.40 & 25.33 & 53.90 & 56.10 & 13.36 & 14.32 & 15.43 & 38.06 & 40.26 & 23.23 & 54.58 & 68.50 & 31.71 & 42.21 \\
\textbf{Llama-Nemo} & 45.00 & 90.80 & 50.00 & 61.50 & 62.80 & 46.07 & 29.52 & 22.86 & 26.20 & 41.46 & 33.84 & 31.12 & 41.48 & 29.41 & 38.80 \\
\textbf{MiMo-RL} & 0.00 & 26.60 & 11.33 & 56.78 & 52.44 & 6.45 & 29.84 & 4.00 & 29.91 & 20.78 & 4.04 & 17.00 & 29.38 & 28.12 & 39.69 \\
\textbf{MiniCPM} & 68.33 & 93.40 & \cellcolor{deepred}95.33 & 34.09 & 32.93 & 63.70 & 30.96 & 62.71 & \cellcolor{lightred}88.06 & 69.93 & 52.02 & 67.22 & 65.36 & 35.72 & 51.30 \\
\textbf{NVIDIA-Nemo} & \cellcolor{lightred}70.00 & 93.60 & \cellcolor{lightred}94.67 & 42.09 & 48.17 & \cellcolor{lightred}65.50 & \cellcolor{lightred}56.80 & 70.71 & 86.48 & \cellcolor{lightred}70.13 & 52.02 & \cellcolor{lightred}71.23 & 74.71 & \cellcolor{lightred}39.02 & 46.75 \\
\textbf{OpenThinker} & 41.67 & 85.60 & 62.67 & 11.81 & 7.32 & 44.17 & 11.04 & 17.29 & 42.50 & 42.06 & 31.82 & 61.55 & 52.08 & 22.81 & 33.44 \\
\textbf{Qwen-Coder} & 1.67 & 65.20 & 67.33 & \cellcolor{deepred}75.98 & \cellcolor{deepred}77.44 & 27.87 & 34.24 & 20.43 & 57.04 & 45.35 & 32.32 & 63.21 & 48.78 & 37.02 & 52.60 \\
\textbf{Qwen3-8B} & \cellcolor{deepred}76.67 & 93.60 & \cellcolor{deepred}95.33 & 54.72 & 62.20 & \cellcolor{deepred}67.68 & 53.92 & \cellcolor{lightred}77.00 & 83.98 & 69.13 & \cellcolor{lightred}57.58 & \cellcolor{deepred}74.46 & \cellcolor{lightred}79.18 & 38.74 & 54.22 \\
\textbf{Cogito-v1} & 1.67 & 51.20 & 56.00 & 51.95 & 69.51 & 17.25 & 42.96 & 24.14 & 69.44 & 57.04 & 35.35 & 61.64 & 65.75 & 37.45 & \cellcolor{lightred}54.95 \\
\textbf{Gemma-2-it} & 1.67 & 48.20 & 54.67 & 62.42 & 64.63 & 17.25 & 34.64 & 9.71 & 60.74 & 52.95 & 27.78 & 64.60 & 63.79 & \cellcolor{deepred}39.74 & 52.60 \\
\textbf{Glm-4-chat} & 1.67 & 49.80 & 49.33 & 62.83 & \cellcolor{lightred}74.39 & 16.87 & 37.28 & 12.43 & 46.94 & 47.05 & 23.23 & 57.28 & 62.14 & \cellcolor{lightred}39.02 & \cellcolor{deepred}55.28 \\
\textbf{Granite-3.3-it} & 6.67 & 69.00 & 59.33 & 36.55 & 51.22 & 14.22 & 32.00 & 21.71 & 31.39 & 44.66 & 31.82 & 62.34 & 61.19 & 36.59 & 50.32 \\
\textbf{Internlm3-it} & 6.67 & 69.00 & 60.00 & 60.47 & 65.85 & 21.23 & 33.84 & 25.86 & 61.94 & 55.14 & 33.84 & 61.99 & 67.24 & 36.87 & 50.97 \\

% \midrule

% \textbf{\textit{Max Expert}} & 78.89 & 95.60 & 95.56 & 75.49 & 76.40 & 67.57 & 58.27 & 79.34 & 89.07 & 71.27 & 59.33 & 74.51 & 79.90 & 39.58 & 55.78 \\
% \textit{\textbf{\textit{Oracle}}} & \textit{87.78} & \textit{98.67} & \textit{100.00}$^\dagger$ & \textit{93.65} & \textit{96.80} & \textit{79.68} & \textit{77.43}$^\dagger$ & \textit{99.91}$^\dagger$ & \textit{99.32}$^\dagger$ & \textit{93.93}$^\dagger$ & \textit{97.00}$^\dagger$ & \textit{88.21} & \textit{98.06}$^\dagger$ & \textit{76.48}$^\dagger$ & \textit{87.62}$^\dagger$ \\
% \midrule
% \textbf{Random Router} & 15.56 & 70.67 & 68.44 & 54.06 & 58.40 & 34.51 & 36.94 & 35.55 & 61.42 & 51.47 & 43.33 & 58.13 & 59.95 & 35.85 & 47.51 \\
% \textbf{RouterDC} & -- & -- & -- & -- & -- & -- & -- & -- & -- & -- & -- & -- & -- & -- & -- \\
% \textbf{EmbedLLM} & 81.11 & 94.67 & 93.78 & 69.42 & 73.2 & 64.16 & 57.76 & 77.35 & 88.03 & 69.2 & 63.00 & 71.11 & 75.81 & 38.37 & 51.62 \\
% \textbf{ModelSAT} & 70.00 & 93.87 & 93.24 & 73.11 & 76.40 & 67.26 & 61.07 & 79.24 & 89.75 & 70.00 & 61.33 & 73.33 & 79.32 & 38.09 & 52.11 \\
% \textbf{GraphRouter} & -- & -- & -- & -- & -- & -- & -- & -- & -- & -- & -- & -- & -- & -- & -- \\
% \textbf{Avengers} & 71.11 & 93.20 & 94.22 & 76.04 & 75.20 & 65.17 & 61.49 & 77.25 & 90.31 & 70.00 & 59.33 & 73.98 & 79.42 & 37.80 & 54.54 \\
\bottomrule
\end{tabular}%
}
\caption{The performance of each model on each dataset under the routing for performance-oriented setting. The \colorbox{deepred}{deep red} and \colorbox{lightred}{light red} markers denote the best and second-best results, respectively.
% \textit{Max Expert} represents the best performance of the ten models on the dataset.  \textit{Oracle} represents the best achievable score by selecting the optimal model per query. $^\dagger$ For datasets that contain multiple-choice questions, the \textit{Oracle} may be overestimated.
}
\label{tab:benchmark-results}
\end{table*}
\begin{table*}[]
\small
\setlength{\tabcolsep}{5pt}
\centering
\renewcommand{\arraystretch}{0.95}
\resizebox{\textwidth}{!}{%
\begin{tabular}{lcccccccccc}
\toprule
\multirow{2}{*}{\textbf{Model}} &
\multicolumn{2}{c}{\textbf{Mathematics}} &
\multicolumn{2}{c}{\textbf{Code}} &
\multicolumn{4}{c}{\textbf{Knowledge}} &
\multicolumn{1}{c}{\textbf{IF}} &
\multicolumn{1}{c}{\textbf{Tool Use}} \\
\cmidrule(lr){2-3}
\cmidrule(lr){4-5}
\cmidrule(lr){6-9}
\cmidrule(lr){10-10}
\cmidrule(lr){11-11}
&
\textbf{AIME} &
\textbf{LMB.} &
\textbf{LCB.} &
\textbf{SWE.} &
\textbf{GPQA} &
\textbf{HLE} &
\textbf{MP.} &
\textbf{SQA.} &
\textbf{AHARD.} &
\textbf{Tau2.} \\

\midrule
\textbf{Claude-v4} & 36.67 & 61.16 & 58.10 & \cellcolor{deepred}34.60 & 68.69 & 4.82 & 83.17 & 15.14 & 54.67 & 49.28 \\
\textbf{DeepSeek-R1} & 85.00 & \cellcolor{lightred}77.69 & \cellcolor{lightred}80.09 & \cellcolor{lightred}28.60 & 79.80 & 15.99 & 84.23 & 27.12 & 63.53 & 37.77 \\
\textbf{DeepSeek-V3} & 43.33 & 61.98 & 66.64 & 25.00 & 59.60 & 3.61 & 78.00 & 27.76 & 60.13 & 7.19 \\
\textbf{DS-V3.1-Tms} & 55.00 & 73.55 & 67.30 & 25.40 & 76.26 & 8.94 & 84.03 & 24.46 & 66.27 & 46.76 \\
\textbf{Gemini-Flash} & 63.33 & 73.55 & 60.00 & 18.20 & 66.16 & 7.78 & 80.93 & 29.82 & 55.53 & 32.37 \\
\textbf{Gemini-Pro} & 85.00 & 50.41 & 79.15 & \cellcolor{deepred}34.60 & \cellcolor{deepred}84.85 & \cellcolor{lightred}21.69 & \cellcolor{lightred}85.67 & \cellcolor{deepred}53.77 & \cellcolor{deepred}77.00 & 43.17 \\
\textbf{GLM-4.6} & 81.67 & 63.64 & 63.13 & 21.80 & 75.25 & 15.52 & 80.33 & 25.68 & 66.40 & \cellcolor{lightred}53.96 \\
\textbf{GPT-5} & \cellcolor{deepred}88.33 & \cellcolor{deepred}79.34 & \cellcolor{deepred}86.45 & 15.80 & \cellcolor{deepred}84.85 & \cellcolor{deepred}26.32 & \cellcolor{deepred}87.37 & 47.90 & 69.73 & \cellcolor{deepred}70.14 \\
\textbf{GPT-5-Chat} & 71.67 & 57.85 & 62.18 & 8.80 & 73.74 & 5.70 & 81.40 & 40.50 & 68.87 & -- \\
\textbf{Intern-S1} & 48.33 & 68.60 & 49.19 & 7.80 & 65.66 & 8.85 & 81.97 & 14.77 & 67.07 & 32.73 \\
\textbf{Kimi-K2} & 65.00 & 71.90 & 67.30 & 23.40 & 73.23 & 6.12 & 80.30 & 29.03 & 71.07 & 48.92 \\
\textbf{Qwen3-235B} & 76.67 & 76.03 & 64.17 & 16.20 & 58.59 & 9.22 & 82.77 & \cellcolor{deepred}53.77 & 73.07 & 40.29 \\
\textbf{Qwen3-Thinking} & \cellcolor{lightred}86.67 & 52.89 & 78.39 & 21.40 & \cellcolor{lightred}80.30 & 7.78 & 79.47 & \cellcolor{lightred}48.27 & \cellcolor{lightred}73.93 & 53.60 \\
% \midrule
% \textbf{Random Router} & 66.67 & 66.49 & 67.00 & 22.00 & 76.00 & 10.06 & 82.87 & 34.48 & 68.13 & 40.95 \\
% \textbf{\textit{Max Expert}} & 87.78 & 81.62 & 85.74 & 35.33 & 84.33 & 26.14 & 88.44 & 54.33 & 77.64 & 69.52 \\
% \textit{\textbf{Oracle}} & \textit{93.33} & \textit{84.86} & \textit{91.67} & \textit{66.00} & \textit{97.00} & \textit{50.06} & \textit{95.20} & \textit{84.95} & \textit{99.91} & \textit{93.57} \\
\bottomrule
\end{tabular}%
}
\caption{The performance of each model on each dataset under the Routing for performance-cost setting. The \colorbox{deepred}{deep red} and \colorbox{lightred}{light red} markers denote the best and second-best results, respectively. Note that GPT-5-Chat has no score on the $\tau^2$-Bench benchmark because this model does not support tool calling.}
\label{tab:llm-benchmark-results-categorized}
\end{table*}
\begin{table*}[t]
\small
\setlength{\tabcolsep}{3pt}
\centering
\renewcommand{\arraystretch}{0.95}
\resizebox{\textwidth}{!}{%
\begin{tabular}{lccccccccccc}
\toprule
\multirow{2}{*}{\textbf{Model}} &
\multicolumn{2}{c}{\textbf{Mathematics}} &
\multicolumn{2}{c}{\textbf{Code}} &
\multicolumn{4}{c}{\textbf{Knowledge}} &
\multicolumn{1}{c}{\textbf{IF}} &
\multicolumn{1}{c}{\textbf{Tool Use}} &
\multirow{2}{*}{\textbf{Total}} \\
\cmidrule(lr){2-3}
\cmidrule(lr){4-5}
\cmidrule(lr){6-9}
\cmidrule(lr){10-10}
\cmidrule(lr){11-11}
&
\textbf{AIME} &
\textbf{LMB.} &
\textbf{LCB.} &
\textbf{SWE.} &
\textbf{GPQA} &
\textbf{HLE} &
\textbf{MP.} &
\textbf{SQA.} &
\textbf{AHARD.} &
\textbf{Tau2.} &
\\
\midrule
\textbf{Claude-v4} & 1.26 & 1.91 & 14.20 & 35.72 & 2.65 & 23.71 & 23.40 & 7.33 & 15.94 & 103.72 & 229.83 \\
\textbf{DeepSeek-R1} & 2.09 & 2.70 & 30.46 & 10.03 & 3.25 & 72.03 & 24.62 & 6.19 & 8.52 & 37.08 & 196.97 \\
\textbf{DeepSeek-V3} & \cellcolor{lightred}0.15 & \cellcolor{deepred}0.17 & 1.53 & 2.31 & \cellcolor{lightred}0.18 & \cellcolor{deepred}1.93 & \cellcolor{lightred}1.76 & 0.48 & \cellcolor{lightred}0.80 & \cellcolor{lightred}6.92 & \cellcolor{deepred}16.22 \\
\textbf{DS-V3.1-Tms} & 0.17 & \cellcolor{lightred}0.19 & \cellcolor{lightred}1.19 & \cellcolor{lightred}2.30 & 0.25 & 4.35 & \cellcolor{deepred}1.67 & 0.80 & 1.08 & 17.38 & 29.39 \\
\textbf{Gemini-Flash} & 2.25 & 1.90 & 6.16 & 3.94 & 2.65 & 71.07 & 10.96 & 0.63 & 5.49 & \cellcolor{deepred}5.80 & 110.86 \\
\textbf{Gemini-Pro} & 9.18 & 10.96 & 142.59 & 80.32 & 16.59 & 277.72 & 117.18 & 41.12 & 34.63 & 38.25 & 768.53 \\
\textbf{GLM-4.6} & 2.79 & 4.02 & 2.32 & 5.10 & 5.35 & 100.59 & 30.60 & 26.54 & 2.51 & 27.61 & 207.41 \\
\textbf{GPT-5} & 4.57 & 4.09 & 54.92 & 24.27 & 7.94 & 137.59 & 36.17 & 57.25 & 27.85 & 61.67 & 416.31 \\
\textbf{GPT-5-Chat} & \cellcolor{deepred}0.08 & 1.57 & 4.02 & 11.52 & 1.85 & 18.41 & 13.44 & 2.37 & 6.97 & -- & 60.22 \\
\textbf{Intern-S1} & 0.46 & 0.50 & 7.10 & 2.86 & 0.81 & 11.09 & 5.59 & 3.78 & 1.99 & 13.28 & 47.45 \\
\textbf{Kimi-K2} & 0.57 & 0.60 & 3.88 & 4.79 & 0.63 & \cellcolor{lightred}2.35 & 3.86 & \cellcolor{lightred}0.46 & 1.70 & 27.90 & 46.75 \\
\textbf{Qwen3-235B} & 0.31 & 0.23 & \cellcolor{deepred}1.04 & \cellcolor{deepred}1.36 & \cellcolor{deepred}0.07 & 4.51 & 1.80 & \cellcolor{deepred}0.20 & \cellcolor{deepred}0.74 & 11.67 & \cellcolor{lightred}21.94 \\
\textbf{Qwen3-Thinking} & 1.13 & 1.04 & 14.74 & 8.88 & 2.42 & 17.61 & 25.64 & 12.04 & 8.02 & 28.44 & 119.95 \\
% \midrule
% \textbf{AVG} & 0.57 & 0.72 & 6.73 & 4.50 & 1.01 & 16.94 & 6.92 & 3.67 & 2.67 & 9.86 & 52.83 \\
\bottomrule
\end{tabular}%
}
\caption{Model inference cost comparison across different datasets (Unit: \$/1M tokens) for performance--cost tradeoff setting. The \colorbox{deepred}{deep red} and \colorbox{lightred}{light red} markers denote the lowest and second-lowest costs, respectively. Note that GPT-5-Chat has no score on the $\tau^2$-Bench benchmark because this model does not support tool calling.}
\label{tab:llm-cost-comparison}
\end{table*}
% \clearpage

\begin{table*}[h!]
\centering
\small
\setlength{\tabcolsep}{5pt}
\begin{tabular}{lcc}
\toprule
\textbf{Model} & \textbf{Abbr.} & \textbf{Params} \\
\midrule
DeepHermes-3-Llama-3-8B-Preview~\cite{teknium2024hermes3technicalreport}& DH-Llama3-it & 8B \\
DeepSeek-R1-0528-Qwen3-8B~\cite{guo2025deepseek} & DS-Qwen3 & 8B \\
DeepSeek-R1-Distill-Qwen-7B~\cite{guo2025deepseek}& DS-Qwen & 7B \\
Fin-R1~\cite{liu2025finr1largelanguagemodel} & Fin-R1 & 7B \\ 
GLM-Z1-9B-0414~\cite{glm2024chatglm} & GLM-Z1 & 9B \\
Intern-S1-mini~\cite{bai2025intern} & Intern-S1-mini & 8B \\
Llama-3.1-8B-Instruct \cite{grattafiori2024llama}& Llama-3.1-it & 8B \\
Llama-3.1-8B-UltraMedical~\cite{zhang2024ultramedical}& UltraMedical & 8B \\
Llama-3.1-Nemotron-Nano-8B-v1 \cite{bercovich2025llamanemotronefficientreasoningmodels}& Llama-Nemo & 8B \\
MiMo-7B-RL-0530~\cite{coreteam2025mimounlockingreasoningpotential} & MiMo-RL & 7B \\
MiniCPM4.1-8B~\cite{team2025minicpm4} & MiniCPM & 8B \\
NVIDIA-Nemotron-Nano-9B-v2~\cite{nano2025efficient}& NVIDIA-Nemo & 9B \\
OpenThinker3-7B~\cite{guha2025openthoughtsdatarecipesreasoning}& OpenThinker & 7B \\
Qwen2.5-Coder-7B-Instruct~\cite{hui2024qwen25codertechnicalreport} & Qwen-Coder & 7B \\
Qwen3-8B~\cite{yang2025qwen3} & Qwen3-8B & 8B \\
Cogito-v1-preview-llama-8B~\cite{deepcogito_cogito_v1_llama8b_2025} & Cogito-v1 & 8B \\
Gemma-2-9b-it~\cite{team2024gemma} & Gemma-2-it & 9B \\
Glm-4-9b-chat~\cite{glm2024chatglm}& Glm-4-chat & 9B \\
Granite-3.3-8b-instruct~\cite{ibm_granite_3_3_8b_instruct_2025} & Granite-3.3-it & 8B \\
Internlm3-8b-instruct \cite{cai2024internlm2} & Internlm3-it & 8B \\
\bottomrule
\end{tabular}
\caption{Model pool for the performance-oriented setting:
open-source models around 7B parameters.}
\label{tab:perf-model-pool}
\end{table*}

\begin{table*}[h!]
\centering
\small
\setlength{\tabcolsep}{5pt}
\begin{tabular}{llcc}
\toprule
\textbf{Model} & \textbf{Abbr.} & \textbf{Input Price} & \textbf{Output Price} \\
\midrule
Claude-sonnet-4\cite{anthropic2025claude4} & Claude-v4 & \$3.00/1M & \$15.00/1M \\
Gemini-2.5-flash\cite{google2025gemini25} & Gemini-Flash & \$0.30/1M & \$2.50/1M \\
Gemini-2.5-pro\cite{google2025gemini25} & Gemini-Pro & \$1.25/1M & \$10.00/1M \\
GPT-5-chat\cite{openai2025gpt5} & GPT-5-Chat & \$1.25/1M & \$10.00/1M \\
GPT-5-medium\cite{openai2025gpt5} & GPT-5 & \$1.25/1M & \$10.00/1M \\
Qwen3-235b-a22b-2507\cite{yang2025qwen3} & Qwen3-235B & \$0.09/1M & \$0.60/1M \\
Qwen3-235b-a22b-thinking-2507\cite{yang2025qwen3} & Qwen3-Thinking & \$0.30/1M & \$2.90/1M \\
Deepseek-v3-0324\cite{liu2024deepseek} & DeepSeek-V3 & \$0.25/1M & \$0.88/1M \\
Deepseek-v3.1-terminus\cite{liu2024deepseek} & DS-V3.1-Tms & \$0.27/1M & \$1.00/1M \\
Deepseek-r1-0528\cite{guo2025deepseek} & DeepSeek-R1 & \$0.50/1M & \$2.15/1M \\
GLM-4.6\cite{zeng2025glm} & GLM-4.6 & \$0.60/1M & \$2.20/1M \\
Kimi-k2-0905\cite{team2025kimi} & Kimi-K2 & \$0.50/1M & \$2.00/1M \\
Intern-s1\cite{bai2025intern} & Intern-S1 & \$0.18/1M & \$0.54/1M \\
\bottomrule
\end{tabular}
\caption{Model pool for the performance–cost setting: flagship models.}
\label{tab:cost-model-pool}
\end{table*}
\begin{table*}[!htbp]
    \small
    \centering
    % \resizebox{\columnwidth}{!}{%
    \begin{tabular}{@{}lclll@{}}   % <-- 新增一列
    \toprule
    \textbf{Dataset} & \textbf{Abbrev.} & \textbf{Category} & \textbf{Metrics} & \textbf{Size} \\ \midrule
    \multicolumn{5}{c}{\textit{\textbf{Routing for Performance-Oriented Datasets}}} \\
    AIME          & AIME   & Mathematics & Accuracy, 0-shot & 60  \\
    MATH500~\cite{lightman2023lets}       & M500.  & Mathematics & Accuracy, 0-shot & 500 \\
    MathBench~\cite{liu2024mathbench}     & MBen.  & Mathematics & Accuracy, 0-shot & 150 \\
    MBPP~\cite{austin2021mbpp}            & MBPP   & Code        & Pass@1, 0-shot   & 974 \\
    HumanEval~\cite{chen2021evaluating}   & HE.    & Code        & Pass@1, 0-shot   & 164 \\
    LiveCodeBench~\cite{jain2024livecodebench} & LCB. & Code     & Pass@1, 0-shot   & 1055 \\ 
    KORBench~\cite{ma2024korbenchbenchmarkinglanguagemodels} & KOR. & Logic & Accuracy, 3-shot & 1250 \\
    Knights and Knaves~\cite{xie2024memorization} & K\&K. & Logic & Accuracy, 0-shot & 700 \\
    BBH~\cite{suzgun2022challenging}      & BBH    & Logic       & Accuracy, 3-shot & 1080  \\
    MMLU-Pro~\cite{wang2024mmlupro}        & MP.    & Knowledge   & Accuracy, 0-shot & 1000  \\
    GPQA~\cite{rein2024gpqa}              & GPQA   & Knowledge   & Accuracy, 0-shot & 198  \\
    FinQA~\cite{chen2021finqa}            & FQA.   & Knowledge   & Accuracy, 0-shot & 1147  \\
    MedQA~\cite{jin2021disease}           & MQA.   & Knowledge   & Accuracy, 0-shot & 1273 \\
    EmoryNLP~\cite{byrkjeland2018ternary} & Emory. & Affective   & Accuracy, 0-shot & 697 \\ 
    MELD~\cite{poria2019meldmultimodalmultipartydataset} & MELD & Affective & Accuracy, 0-shot & 1232 \\ 
    \textbf{Total} & & & & \textbf{11,480} \\ 
    \midrule
    \multicolumn{5}{c}{\textit{\textbf{Routing for Performance--Cost Datasets}}} \\
    AIME          & AIME  & Mathematics & Accuracy, 0-shot & 60  \\
    LiveMathBench~\cite{liu2024livemathbench} & LMB. & Mathematics & Accuracy, 0-shot & 121 \\
    LiveCodeBench~\cite{jain2024livecodebench} & LCB. & Code & Pass@1, 0-shot & 1055 \\ 
    SWE-Bench~\cite{jimenez2023swe} & SWE. & Code & Pass@1, 0-shot & 500 \\
    GPQA~\cite{rein2024gpqa}              & GPQA & Knowledge & Accuracy, 0-shot & 198  \\
    HLE~\cite{phan2025humanity}           & HLE  & Knowledge & LLM as judge\textsuperscript{$\dagger$}, 0-shot & 2158 \\
    MMLU-Pro~\cite{wang2024mmlupro}        & MP.  & Knowledge & Accuracy, 0-shot & 3000  \\
    SimpleQA~\cite{wei2024measuring}      & SQA.  & Knowledge & LLM as judge\textsuperscript{$\dagger$}, 0-shot & 4326 \\
    ArenaHard~\cite{li2024crowdsourced}   & AHARD. & Instruction Following (IF) & LLM as judge\textsuperscript{$\dagger$}, 0-shot & 750 \\
    $\tau^2$-Bench~\cite{barres2025tau}   & TAU2. & Tool Use & Success Rate, 0-shot & 278 \\
    \textbf{Total} & & & & \textbf{12,446} \\ 
    \bottomrule
    \end{tabular}%
    \caption{Detailed information of the datasets (with abbreviations). ``LLM as judge''\textsuperscript{$\dagger$} means that we use an auxiliary LLM to score model outputs with the \textbf{official prompts}: for HLE and SimpleQA we adopt \texttt{o3-mini} as the judge model, and for ArenaHard we adopt \texttt{deepseek-v3-0324}.}
    % }
    \label{tab:benchmarks}
\end{table*}

\begin{table*}[h]
    \small
    \setlength{\tabcolsep}{5pt}
    \centering
    
    % @{\hspace{8pt}} : 保持第二列紧贴第一列 (距离仅 8pt)
    % c               : 第二列设定为居中 (Center)，而不是左对齐
    \begin{tabular}{@{}l @{\hspace{2pt}} ccc@{}}
    \toprule
    \textbf{Embedding Model} & 
    \makebox[1.6cm][c]{\textbf{GraphRouter}} &  % [c] 表示居中
    \makebox[1.25cm]{\textbf{EmbedLLM}} & 
    \makebox[1.2cm]{\textbf{Avengers}} \\
    \midrule
    nli-bert-base           & \makebox[1.2cm][c]{69.60}    & \makebox[1.2cm][c]{70.55} & \makebox[1.2cm][c]{70.43} \\
    all-MiniLM-L6-v2        & \makebox[1.2cm][c]{68.05}    & \makebox[1.2cm][c]{70.95} & \makebox[1.2cm][c]{71.03} \\
    gte-Qwen2-7B-instruct   & \makebox[1.2cm][c]{70.29} & \makebox[1.2cm][c]{71.24} & \makebox[1.2cm][c]{71.94} \\
    \bottomrule
    \end{tabular}
    
    \caption{Performance comparison of GraphRouter, EmbedLLM, and Avengers on different embedding models.}
    \label{tab:embedding-model-comparison}
\end{table*}

\begin{table*}[t!]
\small
\setlength{\tabcolsep}{1pt}
\centering
\resizebox{\textwidth}{!}{%
\begin{tabular}{lcccccccccccccccc}
\toprule
\multirow{2}{*}{\textbf{Model}} &
\multicolumn{3}{c}{\textbf{Mathematics}} &
\multicolumn{3}{c}{\textbf{Code}} &
\multicolumn{3}{c}{\textbf{Logical}} &
\multicolumn{4}{c}{\textbf{Knowledge}} &
\multicolumn{2}{c}{\textbf{Affective}} &
\multirow{2}{*}{\textbf{Avg}} \\
\cmidrule(lr){2-4}
\cmidrule(lr){5-7}
\cmidrule(lr){8-10}
\cmidrule(lr){11-14}
\cmidrule(lr){15-16}
&
\textbf{AIME} &
\textbf{M500.} &
\textbf{MBen.} &
\textbf{MBPP} &
\textbf{HE.} &
\textbf{LCB.} &
\textbf{KOR.} &
\textbf{K\&K.} &
\textbf{BBH} &
\textbf{MP.} &
\textbf{GPQA} &
\textbf{FQA.} &
\textbf{MQA.} &
\textbf{Emory.} &
\textbf{MELD} &
\\
\midrule
\textbf{DH-Llama3-it} & 0.00 & 28.00 & 38.22 & 54.74 & 51.60 & 12.81 & 32.73 & 15.45 & 51.11 & 36.53 & 31.00 & 50.87 & 57.17 & 38.53 & 48.16 & 36.46 \\
\textbf{DS-Qwen3} & \cellcolor{lightred}76.67 & \cellcolor{lightred}94.67 & 92.00 & 36.86 & 25.60 & 64.35 & 55.88 & 60.19 & \cellcolor{deepred}89.07 & \cellcolor{deepred}71.27 & \cellcolor{deepred}59.33 & 70.56 & \cellcolor{deepred}79.90 & 38.53 & 50.97 & 64.39 \\
\textbf{DS-Qwen} & 40.00 & 87.87 & 88.00 & 53.65 & 42.40 & 40.95 & 44.12 & 50.33 & 69.51 & 47.60 & 34.33 & 64.05 & 34.55 & 29.35 & 33.62 & 50.69 \\
\textbf{Fin-R1} & 15.56 & 75.33 & 66.67 & \cellcolor{lightred}68.81 & \cellcolor{deepred}76.40 & 7.51 & 33.69 & 18.29 & 61.91 & 48.93 & 26.33 & 68.58 & 61.73 & 36.81 & 51.46 & 47.87 \\
\textbf{GLM-Z1} & 62.22 & \cellcolor{deepred}95.60 & \cellcolor{lightred}94.22 & 61.57 & 60.80 & 60.13 & 53.48 & 48.15 & 83.83 & 66.33 & 53.67 & 68.41 & 70.26 & 34.13 & 44.43 & 63.82 \\
\textbf{Intern-S1-mini} & \cellcolor{deepred}78.89 & 91.20 & 63.56 & 45.87 & 42.00 & 25.24 & \cellcolor{deepred}58.27 & \cellcolor{deepred}79.34 & 84.07 & 54.80 & 44.00 & 70.96 & 68.27 & 37.19 & 49.78 & 59.56 \\
\textbf{Llama-3.1-it} & 8.89 & 49.47 & 44.44 & 61.84 & 61.60 & 17.03 & 23.95 & 13.46 & 60.56 & 48.67 & 25.00 & 53.60 & 67.33 & 33.75 & 46.86 & 41.10 \\
\textbf{UltraMedical} & 0.00 & 41.20 & 25.33 & 54.74 & 52.40 & 12.68 & 14.37 & 14.88 & 41.42 & 41.47 & 26.00 & 54.36 & 69.63 & 33.65 & 42.27 & 34.96 \\
\textbf{Llama-Nemo} & 46.67 & 90.53 & 47.11 & 61.64 & 60.40 & 44.98 & 30.28 & 23.03 & 26.98 & 41.27 & 34.00 & 30.84 & 40.63 & 29.92 & 38.22 & 43.10 \\
\textbf{MiMo-RL} & 0.00 & 24.67 & 12.89 & 57.06 & 50.40 & 5.99 & 30.01 & 2.65 & 32.84 & 20.73 & 4.00 & 17.19 & 30.31 & 28.11 & 39.08 & 23.73 \\
\textbf{MiniCPM} & 71.11 & 93.60 & \cellcolor{deepred}95.56 & 34.81 & 30.80 & 61.96 & 30.01 & 60.28 & \cellcolor{lightred}87.78 & \cellcolor{lightred}69.13 & 51.33 & 67.71 & 65.86 & 32.98 & 50.81 & 60.25 \\
\textbf{NVIDIA-Nemo} & 71.11 & 93.60 & \cellcolor{deepred}95.56 & 43.62 & 46.80 & \cellcolor{lightred}64.73 & \cellcolor{lightred}56.15 & 70.62 & 87.22 & \cellcolor{lightred}69.13 & 50.33 & \cellcolor{lightred}71.20 & 74.03 & 38.53 & 46.22 & \cellcolor{lightred}65.26 \\
\textbf{OpenThinker} & 40.00 & 85.33 & 63.56 & 12.29 & 8.00 & 43.28 & 11.28 & 17.16 & 44.57 & 42.20 & 28.33 & 61.44 & 53.25 & 24.00 & 32.49 & 37.81 \\
\textbf{Qwen-Coder} & 4.44 & 64.80 & 66.67 & \cellcolor{deepred}75.49 & \cellcolor{deepred}76.40 & 27.63 & 34.91 & 20.76 & 59.14 & 44.47 & 34.67 & 63.47 & 49.27 & \cellcolor{lightred}39.10 & 51.73 & 47.50 \\
\textbf{Qwen3-8B} & 73.33 & 93.47 & 92.89 & 53.92 & 58.40 & \cellcolor{deepred}67.57 & 54.28 & \cellcolor{lightred}75.36 & 83.83 & 67.73 & \cellcolor{lightred}54.00 & \cellcolor{deepred}74.51 & \cellcolor{lightred}77.80 & \cellcolor{deepred}39.58 & 53.51 & \cellcolor{deepred}68.01 \\
\textbf{Cogito-v1} & 0.00 & 50.00 & 53.33 & 51.81 & 69.20 & 15.90 & 42.68 & 23.41 & 71.23 & 57.80 & 38.67 & 63.36 & 64.97 & 38.82 & \cellcolor{lightred}54.05 & 46.35 \\
\textbf{Gemma-2-it} & 2.22 & 46.27 & 52.44 & 61.77 & 67.20 & 16.85 & 35.92 & 9.19 & 62.84 & 53.67 & 28.00 & 65.91 & 64.29 & \cellcolor{deepred}39.58 & 51.24 & 43.83 \\
\textbf{Glm-4-chat} & 3.33 & 48.00 & 47.56 & 64.30 & \cellcolor{lightred}72.00 & 16.72 & 37.41 & 13.18 & 48.02 & 46.60 & 25.67 & 57.32 & 60.99 & 38.53 & \cellcolor{deepred}55.78 & 42.36 \\
\textbf{Granite-3.3-it} & 10.00 & 68.40 & 57.33 & 36.11 & 48.80 & 13.12 & 33.37 & 21.90 & 30.37 & 44.13 & 31.00 & 62.02 & 60.73 & 36.33 & 49.51 & 40.21 \\
\textbf{Internlm3-it} & 8.89 & 69.87 & 56.44 & 60.61 & 62.40 & 20.82 & 34.06 & 26.16 & 63.21 & 56.00 & 33.00 & 61.85 & 66.75 & 36.52 & 50.05 & 47.11 \\

\midrule

\textbf{Dataset Oracle} & 78.89 & 95.60 & 95.56 & 75.49 & 76.40 & 67.57 & 58.27 & 79.34 & 89.07 & 71.27 & 59.33 & 74.51 & 79.90 & 39.58 & 55.78 & 73.10 \\
\textbf{Oracle} & 87.78 & 98.67 & 100.00 & 93.65 & 96.80 & 79.68 & 77.43 & 99.91 & 99.32 & 93.93 & 97.00 & 88.21 & 98.06 & 76.48 & 87.62 & 91.64 \\

\midrule
\textbf{Random Router} & 15.56 & 70.67 & 68.44 & 54.06 & 58.40 & 34.51 & 36.94 & 35.55 & 61.42 & 51.47 & 43.33 & 58.13 & 59.95 & 35.85 & 47.51 & 48.79 \\
\textbf{RouterDC} & 80.00 & 89.47 & 57.78 & 63.00 & 69.60 & 49.72 & 48.21 & 76.87 & 71.67 & 55.80 & 33.34 & 66.84 & 66.33 & 39.71 & 51.57 & 61.33 \\
\textbf{EmbedLLM} & 81.11 & 94.67 & 93.78 & 69.42 & 73.20 & 64.16 & 57.76 & 77.35 & 88.03 & 69.20 & 63.00 & 71.11 & 75.81 & 38.37 & 51.62 & 71.24 \\
\textbf{GraphRouter} & 77.78 & 92.93 & 93.78 & 66.62 & 74.00 & 65.49 & 58.83 & 80.76 & 85.74 & 66.73 & 53.33 & 73.10 & 74.66 & 39.23 & 51.30 & 70.29 \\
\textbf{Model-SAT} & 70.00 & 93.87 & 93.34 & 73.11 & 76.40 & 67.26 & 61.07 & 79.24 & 89.75 & 70.00 & 61.33 & 73.33 & 79.32 & 38.09 & 52.11 & 71.88 \\
\textbf{Avengers} & 71.11 & 93.20 & 94.22 & 76.04 & 75.20 & 65.17 & 61.49 & 77.25 & 90.31 & 70.00 & 59.33 & 73.98 & 79.42 & 37.80 & 54.54 & 71.94 \\
\bottomrule
\end{tabular}%
}
% \caption{The performance of each model on each dataset under the Routing for Performance setting. The \colorbox{deepred}{deep red} and \colorbox{lightred}{light red} markers denote the best and second-best results, respectively.
% \textit{Max Expert} represents the best performance of the ten models on the dataset.  \textit{Oracle} represents the best achievable score by selecting the optimal model per query. $^\dagger$ For datasets that contain multiple-choice questions, the \textit{Oracle} may be overestimated.}
\caption{The performance of all base models and routing methods on each dataset under the routing for performance-oriented setting.
All results are computed on the 30\% test split and averaged over five random seeds used in Experiments.
The \colorbox{deepred}{deep red} and \colorbox{lightred}{light red} markers denote the best and second-best results, respectively.}
\label{tab:benchmark-results-app}
\end{table*}

% Performance of each model and routing method on per dataset under the routing-for-performance setting. Results are evaluated on the 30% test split and averaged over five random train-test splits (seeds 42, 999, 2024, 2025, and 3407). Deep red and light red cells denote the best and second-best scores on each dataset, respectively.
\begin{table*}[!t]
\small
\setlength{\tabcolsep}{1pt}
\centering
\resizebox{\textwidth}{!}{%
\begin{tabular}{lccccccccccc}
\toprule
\multirow{2}{*}{\textbf{Model}} &
\multicolumn{2}{c}{\textbf{Mathematics}} &
\multicolumn{2}{c}{\textbf{Code}} &
\multicolumn{4}{c}{\textbf{Knowledge}} &
\multicolumn{1}{c}{\textbf{IF}} &
\multicolumn{1}{c}{\textbf{Tool Use}} &
\multirow{2}{*}{\textbf{Avg}} \\
\cmidrule(lr){2-3}
\cmidrule(lr){4-5}
\cmidrule(lr){6-9}
\cmidrule(lr){10-10}
\cmidrule(lr){11-11}
&
\textbf{AIME} &
\textbf{LMB.} &
\textbf{LCB.} &
\textbf{SWE.} &
\textbf{GPQA} &
\textbf{HLE} &
\textbf{MP.} &
\textbf{SQA.} &
\textbf{AHARD.} &
\textbf{Tau2.} &

\\
\midrule
\textbf{Claude-v4} & 41.11 & 61.62 & 56.34 & \cellcolor{deepred}35.33 & 71.33 & 4.79 & 83.76 & 15.55 & 54.08 & 48.81 & 47.27 \\
\textbf{DeepSeek-R1} & 84.44 & 77.84 & \cellcolor{lightred}78.42 & 28.40 & 80.67 & 16.46 & 85.11 & 27.70 & 64.81 & 34.76 & 57.86 \\
\textbf{DeepSeek-V3} & 45.56 & 67.57 & 65.17 & 25.20 & 62.67 & 3.83 & 78.87 & 28.18 & 60.59 & 6.43 & 44.41 \\
\textbf{DS-V3.1-Tms} & 56.67 & 76.76 & 65.55 & 25.47 & 77.00 & 9.08 & 84.49 & 24.98 & 66.14 & 45.00 & 53.11 \\
\textbf{Gemini-Flash} & 60.00 & 71.89 & 58.49 & 20.27 & 62.67 & 7.69 & 81.33 & 29.86 & 55.45 & 30.71 & 47.84 \\
\textbf{Gemini-Pro} & 82.22 & 46.49 & 77.54 & \cellcolor{lightred}34.53 & \cellcolor{lightred}84.00 & \cellcolor{lightred}21.43 & \cellcolor{lightred}86.73 & \cellcolor{deepred}54.51 & \cellcolor{deepred}77.17 & 43.33 & 
\cellcolor{lightred}60.80 \\
\textbf{GLM-4.6} & \cellcolor{lightred}86.67 & 65.41 & 61.51 & 21.60 & 76.67 & 15.16 & 81.04 & 26.36 & 66.18 & \cellcolor{lightred}55.48 & 55.61 \\
\textbf{GPT-5} & \cellcolor{deepred}87.78 & \cellcolor{deepred}82.16 & \cellcolor{deepred}85.68 & 16.67 & \cellcolor{deepred}84.33 & \cellcolor{deepred}26.57 & \cellcolor{deepred}88.49 & 48.47 & 69.90 & \cellcolor{deepred}69.52 & 
\cellcolor{deepred}65.96 \\
\textbf{GPT-5-Chat} & 73.33 & 57.84 & 60.32 & 10.13 & 75.00 & 5.74 & 82.13 & 40.69 & 67.02 & - & 52.47 \\
\textbf{Intern-S1} & 50.00 & 69.19 & 47.63 & 7.60 & 67.67 & 9.05 & 82.13 & 14.92 & 67.06 & 30.48 & 44.57 \\
\textbf{Kimi-K2} & 70.00 & 73.51 & 66.12 & 24.40 & 74.67 & 5.93 & 80.47 & 29.24 & 73.45 & 48.10 & 54.59 \\
\textbf{Qwen3-235B} & 80.00 & \cellcolor{lightred}78.92 & 62.78 & 16.40 & 58.67 & 9.48 & 83.80 & \cellcolor{lightred}54.33 & 74.20 & 38.81 & 55.74 \\
\textbf{Qwen3-Thinking} & 82.22 & 53.51 & 76.78 & 21.33 & 80.00 & 7.69 & 80.36 & 49.03 & \cellcolor{lightred}75.09 & 50.48 & 57.65 \\

\midrule
\textbf{Dataset Oracle} & 87.78 & 82.16 & 85.68 & 35.33 & 84.33 & 26.57 & 88.49 & 54.51 & 77.17 & 69.52 & 69.16 \\
\textbf{Oracle} & 93.33 & 84.32 & 91.61 & 66.00 & 97.00 & 50.96 & 95.20 & 85.12 & 99.47 & 93.57 & 85.66 \\

\midrule
\textbf{OpenRouter} & 43.33 & 57.30 & 59.12 & 11.33 & 76.33 & 13.25 & 86.02 & 42.91 & 57.4 & - & 49.67 \\
\midrule

\textbf{Random Router} & 66.67 & 67.57 & 66.69 & 23.33 & 74.00 & 10.65 & 83.16 & 35.16 & 68.03 & 39.29 & 53.45 \\
\textbf{HybridLLM} & 80.00 & 78.91 & 64.11 & 16.26 & 60.67 & 11.05 & 83.73 & 54.24 & 74.26 & 52.38 & 57.56 \\
\textbf{FrugalGPT} & 80.00 & 81.08 & 65.74 & 15.87 & 59.67 & 23.92 & 83.67 & 55.22 & 84.80 & 47.14 & 59.71 \\
\textbf{RouteLLM} & 87.78 & 81.62 & 85.74 & 16.67 & 84.33 & 26.14 & 88.44 & 55.30 & 81.16 & 69.52 & 67.67 \\
\textbf{GraphRouter (CF)} & 80.00 & 78.92 & 63.41 & 16.40 & 58.67 & 9.44 & 83.71 & 54.19 & 85.07 & 40.00 & 56.98 \\
\textbf{GraphRouter (BL)} & 81.11 & 78.92 & 72.68 & 16.40 & 58.67 & 17.65 & 83.73 & 54.88 & 84.62 & 48.33 & 59.70 \\
\textbf{GraphRouter (PF)} & 85.55 & 80.54 & 84.41 & 28.80 & 85.33 & 20.28 & 85.49 & 57.15 & 86.31 & 64.05 & 67.79 \\

\textbf{Avengers ($\alpha$=0.00)} & 45.56 & 66.49 & 62.71 & 16.40 & 58.67 & 3.92 & 79.33 & 54.21 & 68.01 & 0.00 & 45.53 \\
\textbf{Avengers ($\alpha$=0.05)} & 80.00 & 78.38 & 65.55 & 23.60 & 77.00 & 8.33 & 84.20 & 54.30 & 69.42 & 2.86 & 54.36 \\
\textbf{Avengers ($\alpha$=0.10)} & 80.00 & 78.38 & 65.36 & 23.73 & 77.00 & 8.98 & 84.11 & 54.30 & 69.34 & 31.67 & 57.29 \\
\textbf{Avengers ($\alpha$=0.15)} & 80.00 & 78.38 & 65.30 & 23.60 & 77.00 & 8.98 & 84.09 & 54.30 & 69.03 & 34.52 & 57.52 \\
\textbf{Avengers ($\alpha$=0.20)} & 80.00 & 78.38 & 67.70 & 23.60 & 77.00 & 9.07 & 84.22 & 54.35 & 69.07 & 34.52 & 57.79 \\
\textbf{Avengers ($\alpha$=0.25)} & 80.00 & 78.38 & 71.42 & 24.13 & 77.00 & 9.10 & 84.27 & 54.36 & 69.78 & 36.19 & 58.46 \\
\textbf{Avengers ($\alpha$=0.30)} & 80.00 & 78.38 & 75.02 & 24.13 & 77.00 & 9.35 & 84.33 & 54.33 & 69.73 & 44.76 & 59.70 \\
\textbf{Avengers ($\alpha$=0.35)} & 80.00 & 78.38 & 76.28 & 23.87 & 77.00 & 9.75 & 84.62 & 54.36 & 70.00 & 49.52 & 60.38 \\
\textbf{Avengers ($\alpha$=0.40)} & 80.00 & 78.38 & 76.59 & 24.13 & 77.33 & 22.07 & 84.62 & 54.35 & 70.66 & 54.05 & 62.22 \\
\textbf{Avengers ($\alpha$=0.45)} & 80.00 & 78.92 & 76.66 & 24.67 & 78.00 & 25.52 & 85.02 & 54.41 & 70.35 & 56.90 & 63.05 \\
\textbf{Avengers ($\alpha$=0.50)} & 80.00 & 78.38 & 77.22 & 24.67 & 78.67 & 25.96 & 86.49 & 54.44 & 69.65 & 61.67 & 63.71 \\
\textbf{Avengers ($\alpha$=0.55)} & 82.22 & 77.30 & 79.18 & 24.67 & 79.67 & 26.08 & 87.31 & 54.56 & 69.38 & 63.57 & 64.39 \\
\textbf{Avengers ($\alpha$=0.60)} & 82.22 & 77.30 & 81.14 & 28.53 & 79.67 & 26.14 & 87.49 & 54.70 & 70.53 & 65.00 & 65.27 \\
\textbf{Avengers ($\alpha$=0.65)} & 84.44 & 77.30 & 82.46 & 33.07 & 80.00 & 26.17 & 87.44 & 55.15 & 71.11 & 65.00 & 66.21 \\
\textbf{Avengers ($\alpha$=0.70)} & 84.44 & 77.30 & 84.79 & 34.53 & 81.00 & 26.14 & 87.71 & 55.27 & 71.95 & 64.76 & 66.79 \\
\textbf{Avengers ($\alpha$=0.75)} & 83.33 & 78.38 & 85.05 & 34.53 & 82.33 & 26.11 & 87.69 & 56.66 & 71.46 & 65.24 & 67.08 \\
\textbf{Avengers ($\alpha$=0.80)} & 85.56 & 81.08 & 85.74 & 34.53 & 82.33 & 26.11 & 87.84 & 56.73 & 72.39 & 66.43 & 67.88 \\
\textbf{Avengers ($\alpha$=0.85)} & 85.56 & 80.54 & 85.74 & 35.47 & 84.33 & 26.23 & 88.04 & 56.80 & 73.94 & 67.38 & 68.40 \\
\textbf{Avengers ($\alpha$=0.90)} & 85.56 & 80.54 & 85.74 & 34.80 & 84.33 & 26.20 & 88.00 & 57.18 & 74.65 & 67.86 & 68.49 \\
\textbf{Avengers ($\alpha$=0.95)} & 85.56 & 80.54 & 85.80 & 35.47 & 83.67 & 26.30 & 87.98 & 57.20 & 75.22 & 68.10 & 68.58 \\
\textbf{Avengers ($\alpha$=1.00)} & 85.56 & 80.54 & 85.11 & 35.20 & 83.33 & 26.36 & 88.16 & 57.55 & 75.88 & 68.10 & 68.58 \\
\bottomrule
\end{tabular}%
}
% \caption{The performance of each model on each dataset under the Routing for Performance--Cost Tradeoff setting. The \colorbox{deepred}{deep red} and \colorbox{lightred}{light red} markers denote the best and second-best results, respectively.}
\caption{The performance of all base models and routing methods on each dataset under the routing for performance--cost tradeoff setting.
All results are computed on the 30\% test split and averaged over five random seeds used in Experiments.
The \colorbox{deepred}{deep red} and \colorbox{lightred}{light red} markers denote the best and second-best results, respectively. Note that GPT-5-Chat and OpenRouter have no score on the $\tau^2$-Bench benchmark because this model does not support tool calling. For GraphRouter, we report three configurations---Performance First (PF), Balance (BL), and Cost First (CF)---as employed in the original paper. For Avengers-Pro, we report 21 configurations obtained by varying the performance coefficient from 0 to 1 in increments of 0.05 due to space constraints.}
\label{tab:llm-benchmark-results-categorized-app}
\end{table*}
\begin{table*}[!t]
\small
\setlength{\tabcolsep}{1pt}
\centering
\resizebox{\textwidth}{!}{%
\begin{tabular}{lccccccccccc}
\toprule
\multirow{2}{*}{\textbf{Model}} &
\multicolumn{2}{c}{\textbf{Mathematics}} &
\multicolumn{2}{c}{\textbf{Code}} &
\multicolumn{4}{c}{\textbf{Knowledge}} &
\multicolumn{1}{c}{\textbf{IF}} &
\multicolumn{1}{c}{\textbf{Tool Use}} &
\multirow{2}{*}{\textbf{Total}} \\
\cmidrule(lr){2-3}
\cmidrule(lr){4-5}
\cmidrule(lr){6-9}
\cmidrule(lr){10-10}
\cmidrule(lr){11-11}
&
\textbf{AIME} &
\textbf{LMB.} &
\textbf{LCB.} &
\textbf{SWE.} &
\textbf{GPQA} &
\textbf{HLE} &
\textbf{MP.} &
\textbf{SQA.} &
\textbf{AHARD.} &
\textbf{Tau2.} &
\\
\midrule
\textbf{Claude-v4} & 0.37 & 0.57 & 4.35 & 10.89 & 0.85 & 6.98 & 7.04 & 2.21 & 4.88 & 32.43 & 70.56 \\
\textbf{DeepSeek-R1} & 0.62 & 0.83 & 9.50 & 3.05 & 0.97 & 21.31 & 7.46 & 1.85 & 2.53 & 11.49 & 59.60 \\
\textbf{DeepSeek-V3} & \cellcolor{lightred}0.04 & \cellcolor{deepred}0.05 & 0.47 & \cellcolor{lightred}0.70 & \cellcolor{lightred}0.05 & \cellcolor{deepred}0.57 & \cellcolor{lightred}0.53 & \cellcolor{lightred}0.14 & \cellcolor{lightred}0.24 & \cellcolor{lightred}2.20 & \cellcolor{deepred}5.00 \\
\textbf{DS-V3.1-Tms} & 0.05 & \cellcolor{lightred}0.06 & \cellcolor{lightred}0.37 & \cellcolor{lightred}0.70 & 0.07 & 1.27 & \cellcolor{deepred}0.50 & 0.24 & 0.32 & 5.45 & 9.05 \\
\textbf{Gemini-Flash} & 0.70 & 0.65 & 2.08 & 1.21 & 0.78 & 20.56 & 3.35 & 0.17 & 1.57 & \cellcolor{deepred}1.81 & 32.88 \\
\textbf{Gemini-Pro} & 2.80 & 3.34 & 43.08 & 24.09 & 4.84 & 81.62 & 35.50 & 12.31 & 10.18 & 12.13 & 229.89 \\
\textbf{GLM-4.6} & 0.75 & 1.27 & 0.70 & 1.55 & 1.59 & 30.38 & 9.30 & 8.01 & 0.76 & 8.59 & 62.90 \\
\textbf{GPT-5} & 1.36 & 1.43 & 17.65 & 7.31 & 2.30 & 40.27 & 10.87 & 17.09 & 8.32 & 18.99 & 125.60 \\
\textbf{GPT-5-Chat} & \cellcolor{deepred}0.02 & 0.46 & 1.23 & 3.52 & 0.57 & 5.35 & 4.06 & 0.71 & 2.12 & - & 18.04 \\
\textbf{Intern-S1} & 0.14 & 0.16 & 2.18 & 0.85 & 0.24 & 3.20 & 1.68 & 1.12 & 0.59 & 3.88 & 14.02 \\
\textbf{Kimi-K2} & 0.16 & 0.19 & 1.22 & 1.47 & 0.18 & \cellcolor{lightred}0.68 & 1.16 & \cellcolor{lightred}0.14 & 0.52 & 8.65 & 14.37 \\
\textbf{Qwen3-235B} & 0.09 & 0.07 & \cellcolor{deepred}0.32 & \cellcolor{deepred}0.42 & \cellcolor{deepred}0.02 & 1.33 & 0.55 & \cellcolor{deepred}0.06 & \cellcolor{deepred}0.22 & 3.85 & \cellcolor{lightred}6.93 \\
\textbf{Qwen3-Thinking} & 0.30 & 0.32 & 4.45 & 2.68 & 0.69 & 5.39 & 7.70 & 3.61 & 2.39 & 8.90 & 36.43 \\

\midrule
\textbf{Dataset Oracle} & 1.36 & 1.43 & 17.65 & 10.89 & 2.30 & 40.27 & 10.87 & 12.31 & 10.18 & 18.99 & 126.26 \\
\textbf{Oracle} & 0.12 & 0.09 & 3.56 & 1.86 & 0.13 & 9.78 & 0.65 & 1.07 & 0.26 & 5.52 & 23.04 \\

\midrule
\textbf{OpenRouter} & 1.80 & 2.90 & 4.22 & 4.44 & 2.49 & 37.47 & 18.06 & 2.98 & 7.95 & - & 82.31 \\
\midrule
\textbf{Random Router} & 0.27 & 0.76 & 6.90 & 4.80 & 1.03 & 16.35 & 6.98 & 3.81 & 2.56 & 8.72 & 52.17 \\
\textbf{HybridLLM} & 0.14 & 0.29 & 1.71 & 0.96 & 0.20 & 4.68 & 0.84 & 0.08 & 0.92 & 8.82 & 18.65 \\
\textbf{FrugalGPT} & 0.10 & 0.11 & 4.08 & 4.45 & 0.29 & 37.05 & 0.67 & 3.23 & 0.41 & 8.77 & 59.15 \\
\textbf{RouteLLM} & 1.36 & 1.40 & 17.65 & 7.31 & 2.30 & 40.67 & 10.92 & 2.26 & 8.41 & 18.99 & 111.26 \\
\textbf{GraphRouter (CF)} & 0.09 & 0.07 & 0.33 & 0.42 & 0.02 & 1.34 & 0.56 & 0.06 & 0.22 & 3.84 & 6.94 \\
\textbf{GraphRouter (BL)} & 0.13 & 0.07 & 9.57 & 0.49 & 0.02 & 23.40 & 0.56 & 1.22 & 0.42 & 7.41 & 43.29 \\
\textbf{GraphRouter (PF)} & 2.29 & 0.94 & 22.13 & 13.22 & 4.22 & 58.14 & 20.92 & 6.66 & 5.75 & 16.05 & 150.32 \\
\textbf{Avengers ($\alpha$=0.00)} & 0.04 & 0.05 & 0.31 & 0.42 & 0.02 & 0.57 & 0.47 & 0.06 & 0.30 & 0.00 & 2.23 \\
\textbf{Avengers ($\alpha$=0.05)} & 0.09 & 0.07 & 0.37 & 0.71 & 0.07 & 1.16 & 0.51 & 0.06 & 0.30 & 0.06 & 3.40 \\
\textbf{Avengers ($\alpha$=0.10)} & 0.09 & 0.07 & 0.37 & 0.71 & 0.07 & 1.30 & 0.51 & 0.06 & 0.28 & 0.94 & 4.40 \\
\textbf{Avengers ($\alpha$=0.15)} & 0.09 & 0.07 & 0.37 & 0.71 & 0.07 & 1.30 & 0.51 & 0.06 & 0.28 & 1.10 & 4.57 \\
\textbf{Avengers ($\alpha$=0.20)} & 0.09 & 0.07 & 1.15 & 0.71 & 0.07 & 1.32 & 0.52 & 0.06 & 0.28 & 1.10 & 5.37 \\
\textbf{Avengers ($\alpha$=0.25)} & 0.09 & 0.07 & 2.06 & 0.71 & 0.07 & 1.39 & 0.58 & 0.06 & 0.28 & 1.88 & 7.21 \\
\textbf{Avengers ($\alpha$=0.30)} & 0.09 & 0.07 & 3.62 & 0.71 & 0.07 & 1.65 & 0.61 & 0.06 & 0.43 & 3.81 & 11.13 \\
\textbf{Avengers ($\alpha$=0.35)} & 0.09 & 0.07 & 4.32 & 1.04 & 0.07 & 2.44 & 0.74 & 0.10 & 0.61 & 4.95 & 14.44 \\
\textbf{Avengers ($\alpha$=0.40)} & 0.09 & 0.07 & 4.46 & 1.51 & 0.20 & 29.20 & 1.21 & 0.12 & 1.08 & 7.10 & 45.05 \\
\textbf{Avengers ($\alpha$=0.45)} & 0.09 & 0.08 & 5.01 & 1.65 & 0.32 & 37.98 & 1.53 & 0.14 & 1.46 & 7.60 & 55.86 \\
\textbf{Avengers ($\alpha$=0.50)} & 0.42 & 0.49 & 5.75 & 1.93 & 0.64 & 39.14 & 3.29 & 0.19 & 1.98 & 8.93 & 62.75 \\
\textbf{Avengers ($\alpha$=0.55)} & 0.53 & 0.63 & 9.29 & 1.65 & 0.77 & 39.42 & 4.38 & 0.23 & 2.77 & 9.25 & 68.90 \\
\textbf{Avengers ($\alpha$=0.60)} & 0.53 & 0.63 & 11.94 & 5.26 & 0.77 & 39.66 & 4.69 & 0.96 & 3.34 & 10.41 & 78.17 \\
\textbf{Avengers ($\alpha$=0.65)} & 0.62 & 0.76 & 13.53 & 9.32 & 1.40 & 39.84 & 5.80 & 1.53 & 3.91 & 10.63 & 87.34 \\
\textbf{Avengers ($\alpha$=0.70)} & 0.62 & 0.78 & 16.45 & 10.38 & 1.69 & 40.06 & 7.49 & 1.71 & 4.49 & 11.55 & 95.21 \\
\textbf{Avengers ($\alpha$=0.75)} & 0.81 & 0.83 & 16.60 & 10.38 & 2.06 & 40.06 & 7.94 & 2.71 & 4.80 & 13.52 & 99.71 \\
\textbf{Avengers ($\alpha$=0.80)} & 1.19 & 1.20 & 17.40 & 10.38 & 2.06 & 40.67 & 8.59 & 3.54 & 5.20 & 14.19 & 104.41 \\
\textbf{Avengers ($\alpha$=0.85)} & 1.19 & 1.24 & 17.40 & 12.61 & 2.30 & 40.91 & 9.53 & 4.77 & 5.86 & 14.34 & 110.14 \\
\textbf{Avengers ($\alpha$=0.90)} & 1.19 & 1.24 & 17.40 & 15.72 & 2.30 & 41.08 & 9.71 & 5.88 & 6.39 & 15.11 & 116.01 \\
\textbf{Avengers ($\alpha$=0.95)} & 1.19 & 1.24 & 17.27 & 17.18 & 2.91 & 41.43 & 12.17 & 6.60 & 6.69 & 16.94 & 123.62 \\
\textbf{Avengers ($\alpha$=1.00)} & 1.19 & 1.25 & 18.73 & 19.74 & 4.39 & 41.54 & 14.57 & 7.58 & 7.06 & 16.93 & 132.98 \\

\bottomrule
\end{tabular}%
}
% \caption{Model inference cost comparison across different datasets (Unit: \$/1M tokens). The \colorbox{deepred}{deep red} and \colorbox{lightred}{light red} markers denote the lowest and second-lowest costs, respectively.}

\caption{Inference cost comparison for all base models and routing methods across different datasets (Unit: \$/1M tokens).
All results are computed on the 30\% test split and averaged over five random seeds used in Experiments.
The \colorbox{deepred}{deep red} and \colorbox{lightred}{light red} markers denote the lowest and second-lowest costs, respectively. Note that GPT-5-Chat and OpenRouter have no score on the $\tau^2$-Bench benchmark because this model does not support tool calling. For GraphRouter, we report three configurations---Performance First (PF), Balance (BL), and Cost First (CF)---as employed in the original paper. For Avengers-Pro, we report 21 configurations obtained by varying the performance coefficient from 0 to 1 in increments of 0.05 due to space constraints. }
\label{tab:llm-cost-app}
\end{table*}

% \clearpage
\begin{table*}[h!]
    \centering
    \small
    \setlength{\tabcolsep}{5pt}
    \begin{tabular}{ll}
    \toprule
    \textbf{Provider} & \textbf{Models} \\
    \midrule
    OpenAI & gpt-5, gpt-5-mini, gpt-5-nano, gpt-4.1, gpt-4.1-mini, gpt-4.1-nano, gpt-4o-mini, chatgpt-4o-latest \\
    Anthropic & claude-3.5-haiku, claude-opus-4-1, claude-sonnet-4-0, claude-3-7-sonnet-latest \\
    Google & gemini-2.5-pro, gemini-2.5-flash \\
    Mistral & mistral-large-latest, mistral-medium-latest, mistral-small-latest, mistral-nemo \\
    X.AI & grok-3, grok-3-mini, grok-4 \\
    DeepSeek & deepseek-r1 \\
    Meta-Llama & llama-3.1-70b-instruct, llama-3.1-405b-instruct \\
    MistralAI & mixtral-8x22b-instruct \\
    Perplexity & sonar \\
    Cohere & command-r-plus, command-r \\
    \bottomrule
    \end{tabular}
    \caption{Model pool supported by the \texttt{openroute/auto} routing method on OpenRouter. The available models may change over time; we use the model pool as of December 1, 2025.}
    \label{tab:openrouter-model-pool}
\end{table*}

\begin{figure*}[t]
    \centering   
    \includegraphics[width=\linewidth]{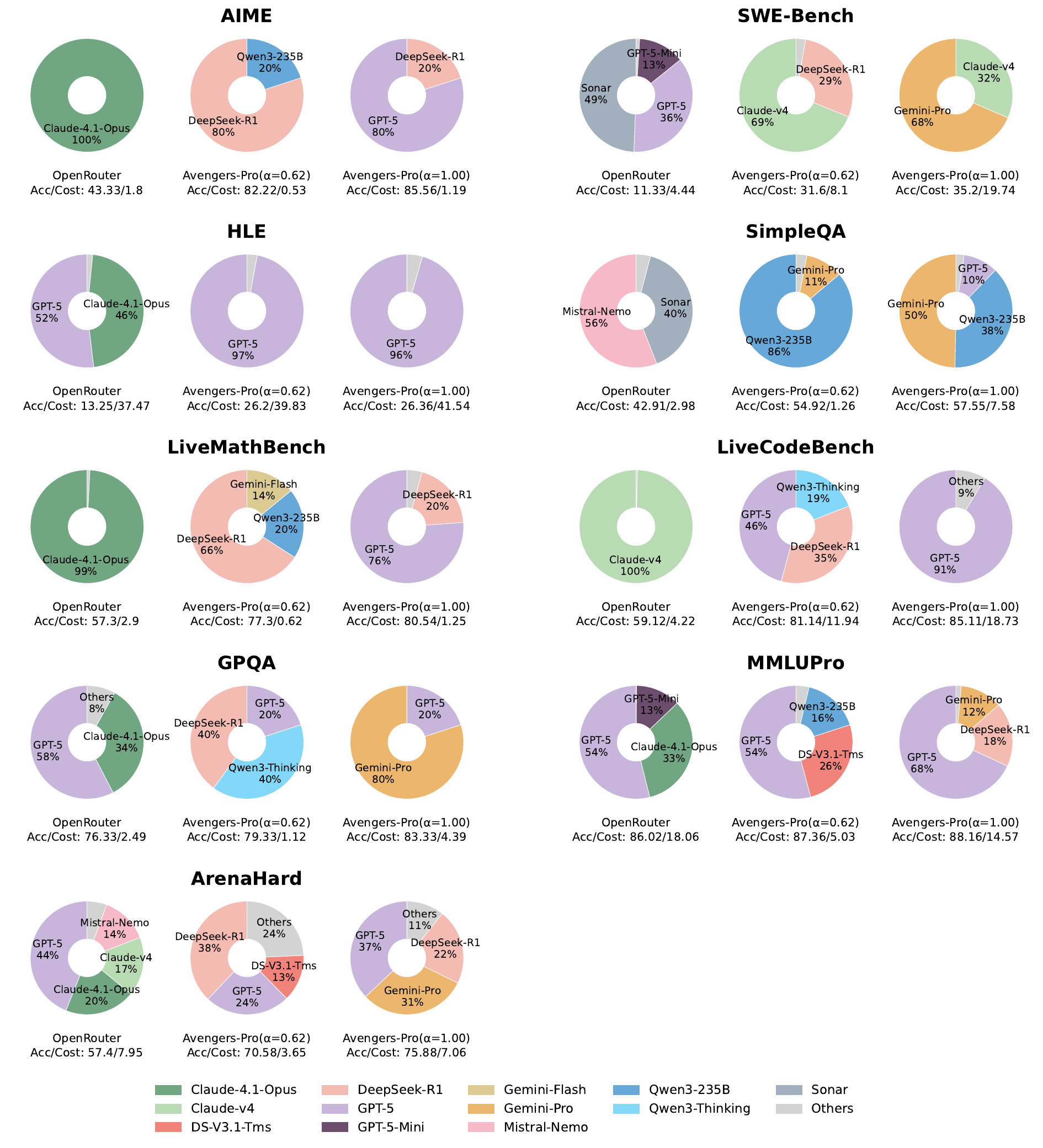}
    \caption{Routing distributions of OpenRouter and Avengers-Pro; Avengers-Pro (cost-matched: $\alpha=0.62$, highest-accuracy: $\alpha=1.00$); Models selected in less than 5\% of queries are grouped into ``Others''.}
    \label{fig:openrouter-2}
\end{figure*}

% \begin{figure*}[t]
%     \centering   
%     \includegraphics[width=\linewidth]{figs/Figure6-how-to-build-model-pool.pdf}
%     \caption{Although the red pool contains consistently weaker single models than the blue pool, it achieves a slightly higher Dataset Oracle and attains comparable overall routing performance (Avengers).}
%     \label{fig:how-to-build-model-pool}
% \end{figure*}

\end{document}